\documentclass[10pt]{article}
\usepackage{amsmath,amssymb,amsthm}
\usepackage[nohead,margin=1.0in]{geometry}

\usepackage{amssymb,amsmath,amsthm}
\usepackage{mathtools}
\mathtoolsset{showonlyrefs}
\usepackage{booktabs}
\usepackage{graphicx,subfigure,epstopdf}
\usepackage[usenames]{color}
\usepackage{url}
\usepackage{algorithm,algorithmic}
\usepackage{dsfont,verbatim}
\usepackage{varwidth}
\usepackage{caption}
\graphicspath{{./pics/}}
\usepackage{enumerate}
\usepackage{tabularx,booktabs}
\usepackage{multirow}

\usepackage{tikz}
\usetikzlibrary{bayesnet}
\allowdisplaybreaks

\newtheorem{example}{Example}[section]

\newtheorem{remark}{Remark}[section]

\newtheorem{proposition}{Proposition}[section]
\numberwithin{equation}{section}

\usetikzlibrary{decorations.pathmorphing} 
\usetikzlibrary{matrix} 
\usetikzlibrary{arrows} 
\usetikzlibrary{calc} 

\tikzstyle{block} = [draw,rectangle,thick,minimum height=1cm,minimum width=3cm, gray, text=black]
\tikzstyle{connector} = [->,thick, gray]
\tikzstyle{line} = [thick, gray]
\tikzstyle{branch} = [circle,inner sep=0pt,minimum size=1.5mm,fill=black,draw=gray]
\tikzstyle{guide} = []
\tikzstyle{snakeline} = [decorate, densely dotted, thick, gray]
\tikzstyle{snakeline*} = [decorate, thick, gray, ->, densely dotted]

\title{Conditional Variational Autoencoder for Learned Image Reconstruction\thanks{A preliminary version of the paper appeared as the arXiv preprint \cite{ZhangJin:2019}. The work of B. Jin is supported by UK EPSRC grant EP/T000864/1}}
\author{Chen Zhang \thanks{Huawei Noah's Ark Lab, Huawei Technologies R\&D, London, UK (\texttt{chenzhang10@huawei.com})}\and Riccardo Barbano\thanks{Department of Computer Science, University College London, Gower Street, London WC1E 6BT, UK (\texttt{riccardo.barbano.19@ucl.ac.uk, b.jin@ucl.ac.uk,bangti.jin@gmail.com})} \and Bangti Jin\footnotemark[3]}

\begin{document}

\maketitle
\begin{abstract}
Learned image reconstruction techniques using deep neural networks have recently gained
popularity, and have delivered promising empirical results. However, most approaches
focus on one single recovery for each observation, and thus neglect the uncertainty information.
In this work, we develop a novel computational framework that approximates the posterior
distribution of the unknown image at each query observation. The proposed framework is very
flexible: It handles implicit noise models and priors, it incorporates the data formation
process (i.e., the forward operator), and the learned reconstructive properties are transferable
between different datasets. Once the network is trained using the conditional variational
autoencoder loss, it provides a computationally efficient sampler for the approximate posterior
distribution via feed-forward propagation, and the summarizing statistics of the generated
samples are used for both point-estimation and uncertainty quantification. We illustrate the
proposed framework with extensive numerical experiments on positron emission tomography (with
both moderate and low count levels) showing that the framework generates high-quality samples
when compared with state-of-the-art methods.
\end{abstract}

\section{Introduction}\label{sec:intro}

Machine learning techniques, predominantly using deep neural networks (DNNs), have received
attention in recent years, and delivered state-of-the-art reconstruction performance on many
classical imaging tasks, including image denoising \cite{zhang2017beyond}, image deblurring
\cite{xu2014deep}, super-resolution \cite{DongLoyHeTang:2014}, and on challenging applications
such as low-dose and sparse-data computed tomography \cite{KangMinYe:2017, ChenZhangWang:2018},
and under-sampled magnetic resonance imaging \cite{HyunKimSeo:2018}, to name just a few.

Most existing works on DNN-based image reconstruction aim at providing one point estimate of
the unknown image of interest for a given (noisy) observation; and that implies, we regard
DNNs as deterministic machineries. Nonetheless, the stochastic nature of practical imaging
problems (e.g., a noisy observation process, or an imprecise prior knowledge) implies that
there actually exists an ensemble of plausible reconstructions, which are still consistent
with the given data (although to various extent). This observation has motivated the need for
developing a fully probabilistic treatment of the reconstruction task in order to assess the
reliability of one specific reconstructed image so as to inform downstream decision making
\cite{kendall2017uncertainties}. Unfortunately, uncertainty / reliability information is not
directly available from most existing DNN-based image reconstruction techniques, and hence there
is an imperative need to develop uncertainty quantification (UQ) frameworks that are capable of
efficiently delivering uncertainty information along with the reconstruction \cite{BarbanoArridgeJinTanno:2022}.

The Bayesian framework provides a systematic yet very flexible way to address many UQ tasks, by
modelling the data and the unknown image probabilistically (i.e., as random variables), and has
been popular for UQ in imaging inverse problems \cite{KaipioSomersalo:2005, Stuart:2010}. Furthermore,
recent advances in Bayesian inference leverage powerful deep generative modelling tools, such as
Wasserstein generative adversarial networks (GANs) \cite{arjovsky2017wasserstein}, and variational
autoencoders (VAEs) \cite{kingma2013auto}. Deep generative models output one sample of the unknown
image via feed-forward propagation, which is computationally feasible, and for multiple samples,
can also be performed in parallel. These techniques hold enormous potentials for quantifying the
uncertainty for image reconstruction; nonetheless, developing rigorous data-driven image reconstruction
techniques within a Bayesian framework remains challenging.

The first challenge stems from the high complexity of posterior distribution of the unknown image (conditioned on a given observation).
The ``conventional'' Bayesian setting often involves an explicit likelihood and the construction of a prior.
In fact, in Bayesian inversion, both likelihood and prior are given explicitly (or hierarchically).
The likelihood is derived from the statistical model of an observation process, under the assumption that both the noise statistics and the underlying physical principles of the imaging modality are well calibrated.
Nonetheless, deriving precise likelihoods can be highly nontrivial (e.g., due to a complex corruption process).
Further, how to stipulate statistically meaningful yet explicit priors is a long-standing open problem in image reconstruction.
In the context of learning-based approaches, one can only implicitly extract prior information in a ``data-driven'' way using deep neural networks from a set of training pairs instead of an explicit posterior distribution (e.g., up to a normalising constant), leading to an approximate posterior, which is ``intrinsically implicit''. Furthermore, the UQ task is substantially complicated by the high-dimensionality of the reconstructed image, which renders the many conventional sampling type techniques not directly applicable.

The second challenge is related to the presence of physical laws. In many practical imaging applications, the data formation process itself is highly complex.
To make matters even worse, the forward operator itself can be implicitly defined, for instance, via a system of differential equations in many medical imaging modalities, such as second-order elliptic differential equation in electrical impedance tomography \cite{Borcea:2002}, or radiative transfer equation in diffuse optical tomography \cite{ArridgeSchotland:2009}.
The forward maps often describe the fundamental physical laws, and can be regarded as ``established'' prior knowledge. Therefore,
it is useful to directly inform DNNs with the underlying physical laws (instead of learning them from the given training data), and how to best use the physical laws represents an important issue in architectural design, which is currently actively researched.

In this work, we develop a novel computational framework for uncertainty-aware learned image reconstruction, using conditional variational autoencoder (cVAE) \cite{sohn2015learning}, implemented via algorithmic unrolling \cite{Monga:2021}.
The framework recurrently refines the (stochastic) reconstruction benefiting directly from the physical laws as well from the samples of a low-dimensional random variable, which is conditionally dependent on the observation.
Furthermore, minimizing the cVAE objective is equivalent to optimizing an upper bound of an expected Kullback-Leibler (KL) divergence (see Proposition \ref{prop:cvae} for a precise statement), and the output samples (i.e., the plausible reconstructions) are actually drawn from an approximate posterior distribution of the unknown image.
In sum, the proposed framework has the following features:
(i) it serves as an efficient sampler from an approximate posterior in the sense of (generalised) variational inference \cite{wainwright2008graphical};
(ii) it is scalable due to its encoder-decoder structure, more specifically to the low-dimensionality of the latent variable. Note that the low-dimensional latent variable in cVAE is introduced to inject stochasticity into the reconstructions, and the heavy-duty reconstruction task is still carried out by the unrolling construction. Thus, the approach does not suffer from the oversmoothing effect typically observed for VAEs.

This work lies within in the context of quantifying uncertainty for learned image reconstruction techniques.
There are mainly two different types of uncertainties: aleatoric \cite{kendall2017uncertainties, lakshminarayanan2017simple} and epistemic \cite{gal2016dropout, kendall2017uncertainties} uncertainties.
See the review \cite{BarbanoArridgeJinTanno:2022} for UQ in medical image analysis, and the review \cite{abdar2020} UQ of DL techniques in general.
We focus on aleatoric uncertainty associated with the reconstructed image.
This arises from the intrinsic stochasticity with the given data, and it is irreducible even if one can collect more data.
It differs from epistemic uncertainty, which often is a byproduct of a Bayesian treatment for neural network, that is, Bayesian neural networks (BNNs) \cite{Graves:2011, Blundell:2015, gal2016dropout, BarbanoZhang:2020, BarbanoKereta:2020}).
BNNs model the uncertainty within network weights, which are then inferred with approximate inference techniques \cite{minka2001expectation, wainwright2008graphical, OsawaSwaroop:2019}.
The corresponding uncertainty arises from the fact that there is an ensemble of weights' configurations that can explain equally well the given training data, since there are generally far more network parameters than the available training data (a.k.a. overparameterization).

The rest of the paper is organised as follows. In Section \ref{sec:related}, we describe the related works. In Section \ref{sec:setting}, we describe the problem formulation. Section \ref{sec:method} provides background information on conditional variational autoencoder, and describes our proposed framework, including the network architecture, and the training and inference phases.
In Section \ref{sec:numerics}, we showcase our framework on an established medical imaging modality -- positron emission tomography (PET) \cite{QiLeahy:2006}, for which uncertainty quantification has long been desired yet still very challenging to achieve \cite{zhang2019expectation, FilipovicBarat:2019, ZhouYuZhang:2020}, and confirm that the generated samples are indeed of high quality in terms of both point estimation and uncertainty quantification, when compared with several state-of-the-art benchmarks.
Finally, in Section \ref{sec:conclusion}, we conclude the paper with additional discussions.

\section{Related works}\label{sec:related}
This work lies at the interface of the following three topics, i.e., learned image reconstruction, uncertainty quantification using generative models, and approximate inference. We review briefly the related works in the following.

Learned reconstruction techniques have received a lot of attention recently. Most of them are learned in a supervised manner \cite{ongie2020deep}. One prominent idea is algorithmic unrolling, i.e., to unfold classical iterative optimization schemes and to replace certain components of the schemes with neural networks, whose parameters are then learned from training data. This construction brings good interpretability into the working of the algorithms, and allows incorporating physical models, thus often delivers high-performant results without resorting to very large training datasets. This was first done in \cite{GregorLecun:2010}, which unrolls the iterative shrinkage thresholding algorithm for sparse coding. In the context of inverse problems, an early work is \cite{putzky2017recurrent}. Currently, there is a large body of works on unrolling; see \cite{Monga:2021} for a recent overview. These approaches focus mostly on point estimates, and ignore the associated uncertainties. This work employs the unrolling idea to construct the backbone network, but also endows the reconstructions with uncertainty estimates using the low-dimensional latent space.

Over the last decade, there have been impressive progresses on UQ of DNNs, especially Bayesian neural networks \cite{Graves:2011,Blundell:2015,gal2016dropout,kendall2017uncertainties,OsawaSwaroop:2019} (see \cite{abdar2020} for an overview). Nonetheless, high-dimensional inference tasks such as image reconstruction remains largely under explored due to the associated computational cost \cite{BarbanoArridgeJinTanno:2022}, and thus UQ of learned image reconstruction techniques is just emerging. For epistemic uncertainty, the most common technique used in image reconstruction is Monte Carlo dropout \cite{gal2016dropout}, due to its low computational cost, but it may sacrifice the accuracy \cite{BarbanoZhang:2020}. Further one may separate the sources of uncertainties into aleatoric and epistemic ones \cite{BarbanoKereta:2020}, if desired. This work addresses the aleatoric uncertainty with a reconstructed image, through approximating an implicitly defined posterior distribution, which differs markedly from existing works on epistemic uncertainty.

In the machine learning community, there is a myriad of different ways to approximate a target distribution using an approximate inference scheme, e.g., variational inference \cite{jordan1999introduction}, variational Gaussian \cite{Opper:2009,arridge2018variational}, expectation propagation \cite{minka2001expectation,zhang2019expectation}, and Laplace approximation \cite{MacKay:2003}. This work provides one way to approximate the distribution using cVAE. Note that to rigorously justify the distribution modelled by DNNs for aleatoric uncertainty, proper Bayesian interpretations of the loss function used in the training of DNNs (in connection with the target posterior distribution) are needed.
The cVAE approach adopted in the present work does admit the interpretation as generalised variational inference (cf. {Proposition} \ref{prop:cvae}). Very recently, a graph version of cVAE was employed for synthesizing cortical morphological changes in \cite{ChaiLiu:2021}. Of course, cVAE represents only one way to construct the approach. Other approaches, e.g., generative adversarial networks \cite{arjovsky2017wasserstein}, also hold potentials, by suitably adapting the corresponding loss function and equipping it with a Bayesian interpretation. For example, the interesting work \cite{HouLamZhang:2019} constructs a sampler by approximating the transport map with DNNs. However, a thorough comparison these different approaches utilizing deep generative models for UQ of learned reconstruction remains missing.

\section{Preliminaries}\label{sec:setting}

In this section we describe the problem setting, and the variational autoencoder (VAE) framework.

\subsection{Problem Formulation}

First we specify the setting of this work, i.e., finite-dimensional linear inverse problems.
Let $x\in\mathbb{R}^n$ be the unknown image of interest, and $y\in\mathbb{R}^m$ be the observational data. The linear forward operator $\mathcal{A}:\mathbb{R}^n\to \mathbb{R}^m$ maps the unknown $x$ to the data $y$.
In practice, the observation $y$ is a noisy version of the exact data $\mathcal{A}x$:
\begin{equation*}
  y = \eta(\mathcal{A}x),
\end{equation*}
where $\eta(\cdot)$ denotes a general corruption process by (possibly unknown type) noise, e.g., Gaussian, Poisson, and Salt and Pepper, Cauchy, uniform or the mixtures thereof.
The image reconstruction task is to recover the unknown ground-truth $x$ from the given noisy observation $y$.
In the Bayesian framework, the corruption process $\eta(\cdot)$ is encoded by a conditional distribution $p^*(y|x)$, and the unknown image $x$ of interest is a random variable with a prior distribution $p^*(x)$.
In the proposed framework, we only require (i) samples from the joint distribution $p^*(x,y):=p^*(y|x)p^*(x)$ (which is proportional to the posterior distribution $p^*(x|y)$, up to a normalizing constant $p(y)=\int_{\mathbb{R}^n} p^*(x,y){\rm d}x$) and (ii) the action of the maps $\mathcal{A}$ and $\mathcal{A}^*$ (the adjoint of $\mathcal{A}$).
These two conditions hold for many practical imaging problems.

Note that DNN-based image reconstruction techniques employ paired training data $\{(x_i,y_i)\}_{i=1}^N$ from the underlying joint distribution
$p^*(x,y)$, and also the associated forward operators, which may differ for each data pair.
Thus the training data tuple takes the form: $(x_i,y_i,\mathcal{A}_i)$, which ideally is an exhaustive representation of the image reconstruction problem.
In particular, a closed-form expression of the posterior distribution $p^*(x|y)$ is not needed, hence the term "implicit" posterior distribution. In practice, one can collect measurements corresponding to samples of $x$ drawn from the prior $p^*(x)$ (i.e., physically derived data), without explicitly knowing the corruption process $\eta(\cdot)$.
However, in many medical imaging applications, a dataset of ordered ground-truth image and observational data pairs is often expensive to acquire at a large volume, if not impossible at all.
Note that the dimension of $(x_i,y_i,\mathcal{A}_i)$ can also vary
from one sample to another, e.g., due to discretization of different resolution levels. The explicit presence of the operator $\mathcal{A}$ is a noticeable difference from standard supervised learning in machine learning.
The main goal of learned image reconstruction with aleatoric UQ is to provide a computational framework that gives an approximation to the posterior distribution $p^*(x|y)$ or the joint distribution $p^*(x,y)$, by suitably training on the given dataset $\{(x_i,y_i,\mathcal{A}_i)\}_{i=1}^N$.

Below we use the notation $h(\cdot)$ to denote a DNN, and use the subscript to distinguish between DNNs. Further, we abuse the subscript for a distribution and a DNN to reparameterize the corresponding random variable.
For instance, $p_\theta(x)$ is a distribution of $x$ and $h_\theta(\cdot)$ is a DNN to reparameterize $x$, both with the parameter vector $\theta$.

\subsection{Variational Inference and Variational Autoencoders}\label{ssec:vae}

Now we describe the basic technique, variational inference (VI) and building block of the proposed framework: variational autoencoders (VAEs).

First we describe the idea of VI for posterior approximation. Let $p_\theta(x|y)$ be an intractable target distribution of interest, where the vector $\theta$ in $p_\theta(x|y)$
contains the hyperparameters of both prior $p_\theta(x)$ and likelihood $p_\theta(y|x)$, such as the prior belief strength (a.k.a. regularisation parameter) and noise precision.
The intractability of the distribution $p_\theta(x|y)$ arises often from its high-dimensionality of the parameter space (i.e., $n$ is large) and non-Gaussian of the distribution nature (e.g., due to the complex data formation process).

To approximate the target distribution $p_\theta(x|y)$, we employ variational inference (VI), which is a popular approximate inference technique in machine learning \cite{jordan1999introduction,wainwright2008graphical}.
It selects the best approximation $q_\phi(x|y)$ from a candidate family
$\mathcal{Q}$ (parameterized by the vector $\phi$, commonly known as the variational parameters) by minimizing a suitable divergence measure for probability distributions.
The family $\mathcal{Q}$ is often taken to be independent Gaussian distributions, which is fully characterised by the mean and (diagonal) covariance. This is commonly known as the mean field approximation.
In VI, the most popular choice of the probability metric is Kullback-Leibler (KL) divergence \cite{KullbackLeibler:1951}.
The KL divergence $D_{\rm KL}(q_\phi(x|y)||p_\theta(x|y))$ of the approximate posterior $q_\phi(x|y)$ from the target posterior $p_\theta(x|y)$ is defined by
\begin{equation*}
	D_{\rm KL}(q_\phi(x|y)||p_\theta(x|y)) = \int q_\phi(x|y)\log\frac{q_\phi(x|y)}{p_\theta(x|y)} \text{d}x.
\end{equation*}
It follows directly from Jensen's inequality that it is always nonnegative and zero if and only if $q_\phi(x|y)=p_\theta(x|y)$
almost everywhere. Since the target $p_\theta(x|y)$ is often only known up to a normalizing constant, the problem
\begin{equation*}
	\min_{q_\phi\in \mathcal{Q}} D_{\rm KL}(q_\phi(x|y)||p_\theta(x|y))
\end{equation*}
is often cast into an equivalent evidence lower bound (ELBO) maximization
\begin{equation}\label{eqn:ELBO}
	\max_\phi \big\{\mathcal{L}(\phi;\theta, y) = \mathbb{E}_{q_\phi(x|y)}[\log p_\theta(y|x)] -D_{\rm KL}(q_\phi(x|y)||p_\theta(x))\big\},
\end{equation}
where the notation $\mathbb{E}_{q}[\cdot]$ denotes taking expectation with respect to the distribution $q$. In the ELBO, the first term enforces data consistency, whereas the second term represents a penalty, which is relative to the prior distribution $p_\theta(x)$.

Solving the optimization problem \eqref{eqn:ELBO} requires evaluating the gradient of the loss $\mathcal{L}$ with respect to the variational parameters $\phi$ (i.e., $\nabla_\phi \mathbb{E}_{q_\phi(x)} [f_\theta(x)]$) for a deterministic function $f_\theta$, parameterized by $\theta$.
The challenge lies in the fact that the integral $\mathbb{E}_{q_\phi(x)}[f_\theta(x)]$ is often not analytically tractable and can only be evaluated by Monte Carlo methods.
The reparameterization trick \cite{kingma2013auto, RezendeMohamed:2014} is useful to overcome the challenges in directly backpropagating the gradients (with respect to $\phi$) through the random variable $x$.
It provides an unbiased estimate of the gradient of the ELBO with respect to the variational parameters $\phi$.
In fact, we assume that the conditional sampling of $x$ depends on $y$ and an easy-to-sample auxiliary variable $\epsilon$ distributed according to $p(\epsilon)$ (e.g., a Gaussian distribution):
\begin{equation*}
	x = g_\phi(y, \epsilon), \quad\text{with}\quad \epsilon\sim p(\epsilon),
\end{equation*}
where $g_\phi(\cdot)$ is a deterministic function (in our case, a DNN). By estimating $\theta$ from the given data $y$, while simultaneously optimizing the variational parameter $\phi$, one obtains the following Monte Carlo estimator of the loss in \eqref{eqn:ELBO}:
\begin{equation*}
	\mathcal{L}(\theta,\phi) = \frac{1}{L}\sum^L_{\ell=1}[\log p_\theta(y_i|x_{(i,\ell)})] - D_{\rm KL}(q_\phi(x|y_i)||p_\theta(x)),
\end{equation*}
where $\{x_{(i,\ell)}\}_{\ell=1}^L$ are $L$ samples generated with $y_i$ and using the variational encoder $q_\phi(x|y)$: $\{\epsilon_\ell\}_{\ell=1}^L$ are sampled from $p(\epsilon)$ and $x_{(i,\ell)}=g_\phi(y_i,z_\ell)$.
The KL term can often be evaluated in closed form since both factors therein are often taken to be Gaussian, otherwise it can also be approximated using Monte Carlo. Note that in the preceding construction, we have allowed the observation data $y$ to vary with the unknown image $x$, in order to accommodate the training data $\{(x_i,y_i)\}_{i=1}^N$. This differs slightly from the standard approximate inference schemes, and it is often referred to as amortized variational inference.

In generative modelling, the above considerations give rise to a formalism very similar to the popular variational autoencoders (VAEs) \cite{kingma2013auto, kingma2019introduction}, with $x$ assuming the role of latent variable, whereas $y$ being the data to be probabilistically modelled; see the remark below for more details on the difference.

VAE is an unsupervised deep generative framework that learns a stochastic mapping between the observed $y$-space and a latent $x$-space. The model learns a joint distribution $p_{\theta}(y, x)$ that factorises as $p_{\theta}(y, x) = p_{\theta}(x)p_{\theta}(y|x)$ with a stochastic decoder $p_{\theta}(y|x)$ and a prior distribution over a low-dimensional latent space $p_{\theta}(x)$. A stochastic encoder $q_{\phi}(x|y)$ (a.k.a. parametric inference model) approximates the intractable posterior $p_{\theta}(x|y)$.
The framework compresses the observed data into a constrained latent distribution (via the encoder) and reconstructs it faithfully (via the decoder). This process is carried out by two neural networks, an encoding network $h_\phi(\cdot)$ with parameter $\phi$ (i.e., the weights and the biases of the network) also called variational parameters, and a decoding network $h_\theta(\cdot)$ with parameter $\theta$.

In practice, VAEs often do not use the encoding network to model the parameterization function $g_\phi(\cdot)$ in an end-to-end way.
VAEs take, instead, the observation $y$ and outputs the coefficients to reparameterize the image $x$. For example, for a multivariate Gaussian $q_\phi(x|y)$ the decoding network can output the mean $\mu$ and Cholesky factor $L$ of the covariance $\Sigma=LL^\top$:
\begin{align*}
    \left(\mu, L\right) &=h_{\phi}(y),\quad
    q_\phi(x|y) =\mathcal{N} (x|\mu,\Sigma).
\end{align*}
We can generate samples from $q_\phi(x|y)$ by sampling $\epsilon$ from the standard
Gaussian $p(\epsilon)= \mathcal{N}(\epsilon|0,I)$, followed by an affine transformation
$ x=\mu+L\epsilon $.

VAE allows performing simultaneously VI (with respect to $\phi$) and model selection
(with respect to $\theta$), and the resulting VAE objective is given by
\begin{equation*}
	\max_{\phi,\theta} \left\{\mathcal{L}_{\rm VAE}(\theta, \phi; y) = \mathbb{E}_{q_\phi(x|y)}[\log p_\theta(y|x)] -D_{\rm KL}(q_\phi(x|y)||p_\theta(x))\right\}.
\end{equation*}
In practice, one may employ an identity variance Gaussian with the mean being the decoder's
output as the likelihood $p_{\theta}(y|x)$, and a standard Gaussian distribution as the prior distribution $p_{\theta}(x)$.

\begin{remark}
Note that the original formalism of VAE \cite{kingma2013auto} is slightly different from the
above. We briefly recall its derivation for the convenience of the readers. Given a dataset
$\{y_i\}_{i=1}^N$ from an unknown probability function $p(z)$ and a multivariate latent
encoding vector $z$, the objective of VAE is to model the data $y$ as a distribution
$p_{\theta }(y)$, i.e.,
$$p_{\theta }(y)= \int p_{\theta }(y|z)p_\theta(z)\,{\rm d}{z},$$
where $\theta$ represents the network parameters. Thus, in VAE, if we assume $  p_{\theta }
({y|z} )$ is a Gaussian distribution, then $p_{\theta }(y)$ is a mixture of Gaussians. To
solve for $\theta$ in a learning paradigm, one constructs an approximation $q_{\phi }({z|y})
\approx p_{\theta }({z|y} )$ by means of VI with variational parameters $\phi$. Repeating the
preceding discussion on VI directly yields the following standard VAE loss
\begin{equation*}
\mathcal{L}(\theta,\phi) = \mathbb{E}_{q_\phi(z|y)}\log p_{\theta }(y|z)-D_{\rm KL}(q_{\phi }(z|y)|| p_{\theta }(z)).
\end{equation*}
The problem then reduces to an autoencoder formalism: the conditional likelihood
distribution $ p_{\theta }(y|z)$ is the probabilistic decoder, and the approximation
$ q_{\phi }(z|y)$ serves as the probabilistic encoder. Hence, the goal of VAE is to
represent the given unlabelled data $\{y_i\}_{i=1}^N $ and to generate new data (from
the latent variable $z$), which differs markedly from the task in learned image
reconstruction, for which the image (represented as latent variable) is of the main
object of interest (and often of very high dimensionality, larger than that of $y$).
Besides, the problem of modelling $p_{\theta }(y|z)$ and $p_\theta(z)$ in the VAE
framework is usually unidentifiable, in the sense that there may exist many different
$(p_{\theta }(y|z), p_\theta(z))$ pairs that admit the same marginal distribution
$p_{\theta }(y)$ \cite{khemakhem2020variational}. Thus the associated modelled approximate
posterior $q_\phi(z|y)$ is not unique.
\end{remark}

\section{Proposed framework}\label{sec:method}

We develop a computational UQ framework that learns a map from the observation $y$
to a distribution, denoted by $p_\phi(x|y)$, which approximates the true posterior
distribution $p^*(x|y)$. The map is modelled with a recurrent  network and a
probabilistic encoder therein allows for diverse reconstruction samples, and hence
it facilitates the quantification of the associated aleatoric uncertainty.

\subsection{Conditional VAE as approximate inference}

Note that a direct application of VAEs to image reconstruction is problematic:
VAEs are unsupervised generative machineries, and would only use noisy observations
$y$, but not the ground-truth image $x$ during the training process. To circumvent
the issue, we employ the cVAE loss \cite{sohn2015learning,WalkerDoerschGuptaHebert:2016},
a conditional variant of VAE:
\begin{equation}\label{eqn:cvae}
	\max_{\phi, \theta} \big\{\mathcal{L}_{\rm cVAE}(\theta, \phi; x, y) = \mathbb{E}_{q_\theta(z|x, y)}[\log p_{\phi_2}(x|y, z)] - D_{\rm KL}(q_\theta(z|x, y)||p_{\phi_1}(z|y))\big\}.
\end{equation}

In cVAEs, there are three distributions: a teacher encoder $q_\theta(z|x,y)$, a student encoder $p_{\phi_1}(z|y)$
and a conditional decoder $p_{\phi_2}(x|y,z)$. The vector $\phi=(\phi_1, \phi_2)$ assembles the parameters of
$p_{\phi_1}(z|y)$ and $p_{\phi_2}(x|y,z) $. The approximate posterior $p_\phi(x|y)$ obtained by cVAE is given by
\begin{equation*}
    p_\phi(x|y)=\int p_{\phi_2}(x|y,z)p_{\phi_1}(z|y)\mathrm{d}z.
\end{equation*}
The cVAE loss admits the following approximate inference
interpretation in a generalised sense. Note that $J^*$ is a functional of the variational distribution $p_\phi(x|y)$ and other auxiliary distributions involving $z$.
\begin{proposition}\label{prop:cvae}
Minimizing the loss $\mathcal{L}_{\rm cVAE}(\theta, \phi; x, y)$ in \eqref{eqn:cvae} expected on the training data distribution $p^*(x,y)$ is equivalent to optimizing an upper bound of the expected $\mathrm{KL}$ divergence
\begin{equation*}
	J^*(p_\phi(x|y)) = \mathbb{E}_{p^*(y)}[D_{\rm KL}(p^*(x|y)||p_\phi(x|y))].
\end{equation*}
\end{proposition}
\begin{proof}
By the definition of KL divergence and Fubini theorem,
\begin{align}
J^*(p_\phi(x|y)) &= \mathbb{E}_{p^*(y)}[D_{\rm KL}(p^*(x|y)||p_\phi(x|y))]\nonumber\\
			&= \int p^*(y)\int p^*(x|y)\log\frac{p^*(x|y)}{p_\phi(x|y)}\text{d}x\text{d}y\nonumber\\
			&= \int p^*(x,y) [\log p^*(x|y) - \log p_\phi(x|y)] \text{d}(x,y)\nonumber\\
			&= \mathbb{E}_{p^*(x,y)}[\log p^*(x|y)] + \mathbb{E}_{p^*(x,y)}[-\log p_\phi(x|y)].\label{eq:ori}
\end{align}
Next we derive a lower bound for $\log p_\phi(x|y)$ of the conditional distribution $p_\phi(x|y)$ using the standard procedure.
Since $p_\phi(x,z|y)=p_\phi(z|x,y)p_\phi(x|y)$, we have
$$p_\phi(x|y)=\frac{p_\phi(x,z|y)}{p_\phi(z|x,y)},$$
and consequently, for any distribution $q(z|x,y)$, there holds
\begin{equation*}
\begin{split}
\log p_\phi(x|y) &= \int q(z|x,y)\log p_\phi(x|y) \text{d}z
			= \int q(z|x,y) \log\frac{p_\phi(x|y)p_\phi(z|x,y)}{p_\phi(z|x,y)} \text{d}z\\
			&= \int q(z|x,y) \log\frac{p_\phi(x,z|y)}{p_\phi(z|x,y)} \text{d}z
		= \int q(z|x,y) \log\frac{p_\phi(x,z|y)}{q(z|x,y)}\frac{q(z|x,y)}{p_\phi(z|x,y)} \text{d}z\\
			&= \int q(z|x,y) \log\frac{p_\phi(x,z|y)}{q(z|x,y)} \text{d}z + \int q(z|x,y) \log\frac{q(z|x,y)}{p_\phi(z|x,y)} \text{d}z.
\end{split}
\end{equation*}
By the nonnegativity of the KL divergence, the second term is nonpositive, and then using the splitting $p_\phi(x,z|y)=p_{\phi_2}(x|y,z)p_{\phi_1}(z|y)$, we deduce
\begin{align*}
\log p_\phi(x|y)	&\ge \int q(z|x,y) \log\frac{p_\phi(x,z|y)}{q(z|x,y)} \text{d}z
			= \int q(z|x,y) \log\frac{p_{\phi_1}(z|y)p_{\phi_2}(x|y,z)}{q(z|x,y)} \text{d}z\\
			&= \int q(z|x,y) \log\frac{p_{\phi_1}(z|y)}{q(z|x,y)} \text{d}z + \int q(z|x,y)\log p_{\phi_2}(x|y,z) \text{d}z\\
			&= -\text{KL}(q(z|x,y)||p_{\phi_1}(z|y)) + \mathbb{E}_{z\sim q(z|x,y)}[\log p_{\phi_2}(x|y,z)],
\end{align*}
i.e.,
\begin{equation}\label{eq:ineq}
	-\log p_\phi(x|y) \le D_{\text{KL}}(q(z|x,y)||p_{\phi_1}(z|y)) + \mathbb{E}_{z\sim q(z|x,y)}[-\log p_{\phi_2}(x|y,z)].
\end{equation}
Taking expectation of \eqref{eq:ineq} with respect to $p^*(x,y)$ and then substituting it into identity \eqref{eq:ori} yield
\begin{align*}
  J^*(p(x|y)) &\leq  \mathbb{E}_{p^*(x,y)}[\log p^*(x|y)] + \mathbb{E}_{p^*(x,y)}[D_{\text{KL}}(q(z|x,y)||p_{\phi_1}(z|y))]\\
     &\quad + \mathbb{E}_{p^*(x,y)}[\mathbb{E}_{z\sim q(z|x,y)}[-\log p_{\phi_2}(x|y,z)]].
\end{align*}
Since the term $\mathbb{E}_{p^*(x,y)}[\log p^*(x|y)]$ is independent of the variational distribution $p_\phi(x|y)$ and other auxiliary distributions involving $z$, minimizing the cVAE loss in \eqref{eqn:cvae} expected on the training data distribution $p^*(x,y)$ is equivalent to minimizing an upper bound of $J^*(p_\phi(x|y))$. This shows the desired assertion.
\end{proof}

In view of Proposition \ref{prop:cvae}, cVAEs can indeed learn an optimal map from the observation $y$ to an approximate posterior distribution $p_\phi(x|y)$ in the sense of minimizing an expected loss of the KL divergence. This interpretation underpins the validity of the procedure for quantifying aleatoric uncertainty. The minimizer is indeed an approximation to the target posterior distribution $p^*(x|y)$, constructed by a generalised variational inference principle.

\begin{example}
We briefly validate Proposition \ref{prop:cvae}, which states that the cVAE framework can approximate the ground-truth distribution $p^*(x|y)$ in the sense of generalised variational inference. Note that it is notoriously challenging to numerically verify the accuracy of any approximate inference scheme for high-dimensional distributions. Nonetheless, in the low-dimensional case, the target distribution can be efficiently sampled by Markov chain Monte Carlo \cite{Liu:2001}.
To illustrate this, we take the ground truth distribution $p^*(x|y)$ to be a two-dimensional multi-modal distribution, which consists of the mixture
of seven Gaussians, shown in Figure \ref{fig:toy}(a), and approximate it by cVAE (with teacher encoder, student encoder and decoder modelled by different three-layer neural networks and ReLu as the nonlinear activation function). We clearly observe from Figure \ref{fig:toy}(b) that the approximation $p_\phi(x|y)$ is fairly accurate, and can
capture well the multi-modality of the distribution, thereby verifying Proposition \ref{prop:cvae}.

\begin{figure}[hbt!]
\centering
\setlength{\tabcolsep}{0pt}
\begin{tabular}{cc}
\includegraphics[trim={.5cm .5cm .5cm .5cm}, clip, width=.45\textwidth]{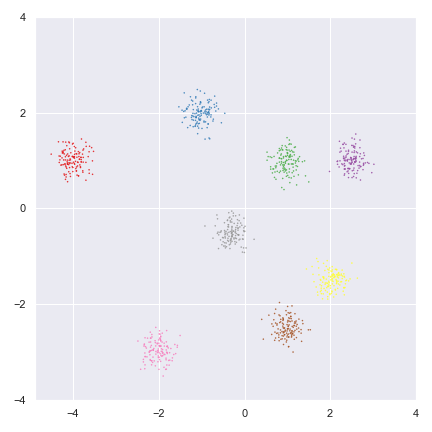} & \includegraphics[trim={.5cm .5cm .5cm .5cm}, clip,width=.45\textwidth]{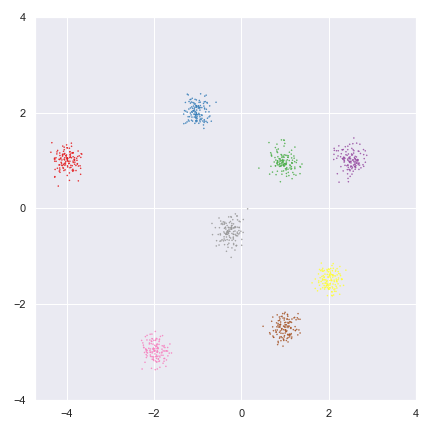}\\
(a) ground-truth & (b) cVAE approximation
\end{tabular}
\caption{Validation of the cVAE for a toy two-dimensional distribution, Gaussian mixture with eight components.\label{fig:toy}}
\end{figure}
\end{example}

We model the conditional encoder $p_{\phi_2}(x|y,z)$ by a mean-field Gaussian with covariance $\beta I$, where $\beta>0$
is a hyperparameter.
The DNN $h_{\phi_2}$ with parameter $\phi_2$ only outputs the mean of $p_{\phi_2}(x|y,z)$,
and on a mini-batch $\{(x_i, y_i)\}^M_{i=1}$ the objective function $\hat{\mathcal{L}}_{\rm cVAE}(\phi,\theta)$ is given by:
\begin{equation*}
	\mathcal{\hat L}_{\rm cVAE}(\phi, \theta) = -\frac{1}{2M}\sum^M_{i=1}\frac{1}{L}\sum_{\ell=1}^L\|x_i-\hat{x}_{(i,\ell)}
\|^2-\frac{\beta}{M}\sum^M_{i=1}D_{\rm KL}(q_\theta(z|x_i,y_i)||p_{\phi_1}(z|y_i)),
\end{equation*}
where $\hat{x}_{(i,\ell)}$ is the mean of the distribution $p_{\phi_2}(x|y_i, z_{i, \ell})$ and $\{z_{i, \ell}\}_{\ell=1}^L$ are $L$ samples
drawn from the distribution $q_\theta(z|x_i, y_i)$, with the default choice $L=1$.
Note that, for special choices of $q_\theta(z|x,y)$ and $p_{\phi_1}(z|y)$ the KL divergence term may be evaluated analytically, and can be used, if available. The gradient of the loss $\mathcal{\hat L}_{\rm cVAE}^L(\phi,\theta)$ is then evaluated by the reparameterization trick.

\begin{remark}
In VAEs, the approximate posterior of the image $x$ (i.e., the latent variable) is modelled by $q_\phi(x|y)$, whereas in cVAEs, it is modelled by
$p_\phi(x|y)=\int p_{\phi_2}(x|y,z)p_{\phi_1}(z|y)\mathrm{d}z.$
In both VAEs and cVAEs, the stochasticity of $x$ comes
from $z$: in VAEs, $z$ is independent on the observation $y$, whereas in cVAEs, $z$ is dependent on $y$. Since the distribution
of $z$ is learned, it can be more flexible than that in VAEs. Thus, even if $p_{\phi_2}(x|y,z)$ is chosen to be simple distributions (e.g., Gaussian distributions) $p_{\phi_1}(x|y)$ can still model a broad family of distributions for continuous unobservable $x$ due to the presence of $p_{\phi_1}(z|y)$, in a manner similar to scale mixture of Gaussians.
\end{remark}

\subsection{cVAE for learned reconstruction}

\begin{figure}[htbp]
 \begin{minipage}{0.25\linewidth}
  \centering
  \begin{tikzpicture}
  \node[obs]                                (y) {$y$};
  \node[const,  left=of y,  xshift=-0cm]    (A) {$\mathcal{A}$};
  \node[latent, below=of y, xshift=0cm]     (x) {$x$};
  \node[latent, below=of y, xshift=1.2cm]   (z) {$z$};
  \edge {y} {x} ; %
  \edge {z} {x} ;
  \edge {y} {z} ;
  \edge[dotted] {A} {x};
  \end{tikzpicture}
  \caption*{(a) Graphical model. Shaded and non-shaded nodes denote observations and hidden variables, respectively, and solid and dotted arrows denote probabilistic dependencies and explicit incorporations, respectively.}
  \end{minipage}%
 \,\,\,\,\,\begin{minipage}{0.70\linewidth}
  \centering
  \includegraphics[width=0.8\textwidth]{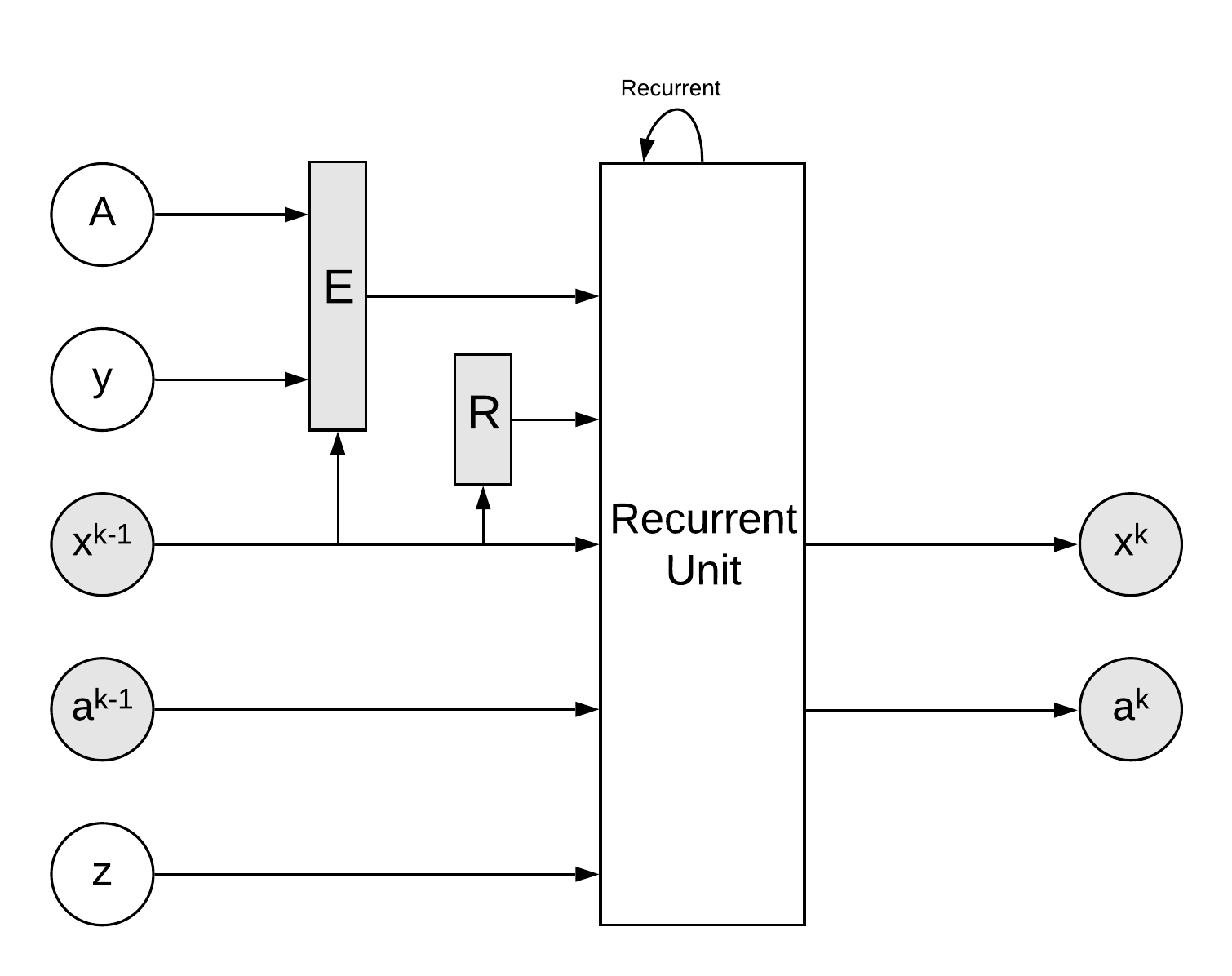}
  \caption*{(b) Recurrent network $h_{\phi_2}$. Shaded and non-shaded circles denote variables for updating and fixed ones, respectively, and shaded rectangles denote functional input. For each fixed $z$, only the values of shaded objects change.}
  \end{minipage}
\caption{The graphical model and recurrent network of cVAE.}	
\label{fig:pdin_gm}
\end{figure}

Probabilistic modelling consists of a learning principle given by a loss function with a proper
probabilistic interpretation, and a graphical model describing probabilistic dependency between variables.
In the proposed framework, we employ a cVAE type loss function:
\begin{equation}\label{eqn:cvae-pdin}
  \begin{aligned}
	\max_{\phi, \theta} \big\{\mathcal{L}(\theta, \phi; x, y, \mathcal{A}) &= \mathbb{E}_{q_\theta(z|x, y, \mathcal{A})}[\log p_\phi(x|y, z,
\mathcal{A})] - D_{\rm KL}(q_\theta(z|x, y, \mathcal{A})||p_\phi(z|y))\big\}.
  \end{aligned}
\end{equation}
Its difference from the standard cVAE loss \eqref{eqn:cvae} is that \eqref{eqn:cvae-pdin} also includes the
forward map $\mathcal{A}$ (and its adjoint $\mathcal{A}^*$) as a part of the training data. Here $\mathcal{A}$ may have different realizations (e.g., corresponding to different levels of discretization) with varying dimensions. Nonetheless, it is viewed as a deterministic variable.
The modelled approximate posterior distribution $p_\phi(x|y)$ is then given by
\begin{equation*}
	p_\phi(x|y)=\int p_{\phi}(x|y,z,\mathcal{A})p_{\phi}(z|y)\text{d}z.
\end{equation*}
The graphical model is given in Figure \ref{fig:pdin_gm}(a).
The learning algorithm that learns a conditional sampler, in a manner similar to a random number generator (RNG) for a given probability distribution.
Note that for a RNG, different runs lead to different samples, but with a fixed random seed, the path is the same for different runs.
The auxiliary (low-dimensional) latent variable $z$ (conditionally dependent on the observation $y$)
is an analogue of the random seed in the RNG, and is introduced into the deterministic recurrent process (modelled by a network) to diversify the reconstruction samples. In particular, for a fixed $z$, the recursion process inputs the sample initialization $x^0$ and applies a recurrent refining step based on suitable sample quality measures and the information encoded in the variable $z$.

Three DNNs are employed to model the distributions in $\mathcal{L}(\theta, \phi; x, y, \mathcal{A})$
and to carry out the conditional sampling process, including one for the recursive process (i.e., recurrent unit modelled by a neural network) and two for probabilistic encoders (i.e., teacher and student encoders).
The observation $y$ and forward map $\mathcal{A}$ constitute their inputs: $y$ is fed into the two probabilistic encoders and each recurrence of the recurrent unit; $\mathcal{A}$ is fed into each recurrence of the recurrent unit and the teacher encoder. Below we explain how the three networks work separately.

The recurrent component is the (deep) network $h_{\phi_2}(\cdot)$; see Figure \ref{fig:pdin_gm}(b).
The network begins with an initial guess $x^0$ (default: back-projected data $\mathcal{A}^*y$) and outputs $x^K$ after $K$ iterations as the mean of $p_\phi(x|y, z, \mathcal{A})$, following the established idea of algorithmic unrolling \cite{putzky2017recurrent,Monga:2021}, which mimics the iterations from standard optimization algorithms (e.g., (proximal) gradient descent, alternating direction method of multipliers (ADMM), and primal-dual hybrid gradient).
At the $k$-th recursion, the network $h_{\phi_2}$ takes one sample $x^{k-1}$ to refine and outputs an improved sample $x^k$. To incorporate the forward map $\mathcal{A}$ and the observation $y$, we employ a functional $E(\mathcal{A},y,x^{k-1})$.
In the lens of variational regularization \cite{ItoJin:2015}, $E(\mathcal{A},y,x^{k-1})$ measures how well $x^{k-1}$ can explain the data $y$. To indicate how well $x^{k-1}$ fulfils the prior knowledge (e.g., sparsity and smoothness) we use the penalty $R(x^{k-1})$ as a part of the input. For the sample quality measure $E(\mathcal{A},y,x^{k-1})$ and the penalty $R(x^{k-1})$, we use $||y-\mathcal{A}(x^{k-1}) ||^2$ (or its gradient), and $||x^{k-1}||^2_2$ or $|x^{k-1}|_{\rm TV}$ (total variation semi-norm), respectively.
Besides the latest iterate $x^{k-1}$ and the quality measures $E$ and $R$, the network $h_{\phi_2}(\cdot)$ also takes a memory variable $a^{k-1}$ and an auxiliary variable $z$.
The memory variable $a^{k-1}$ plays the role of momentum in gradient type methods, and is to retain long-term information between recursions.
The overall procedure resembles an unrolled momentum accelerated gradient descent. The auxiliary random variable $z$ is low-dimensional and to injects randomness into the iteration procedure. Since both $x^{k-1}$ and $x^k$ belong to the image space $\mathbb{R}^n$, we
adopt CNNs without pooling layers to model the recurrent unit $h_{\phi_2}$.
Different inputs of $h_{\phi_2}(\cdot)$ are concatenated
along the channel axis, and the outputs of $h_{\phi_2}(\cdot)$, that is, the update $\delta x^k$ with $x^k=x^{k-1}+\delta x^k$ and the updated memory $a^k$, are also concatenated.

\begin{remark}
At each step, the recurrent unit (neural network $h_{\phi_2}$) reuses the observation data $y$ and the map $\mathcal{A}$ for refinement, and the overall
procedure differs from the deterministic mapping that serves as a post-processing step of back-projection.
The latter is an end-to-end mapping that takes the back-projected data and outputs a refinement, but the proposed approach employs the current sample and quality measures and then decides the refinement strategy.
\end{remark}

\begin{figure*}[htp!]
\centering
\begin{tabular}{cc}
\includegraphics[width=0.3\textwidth]{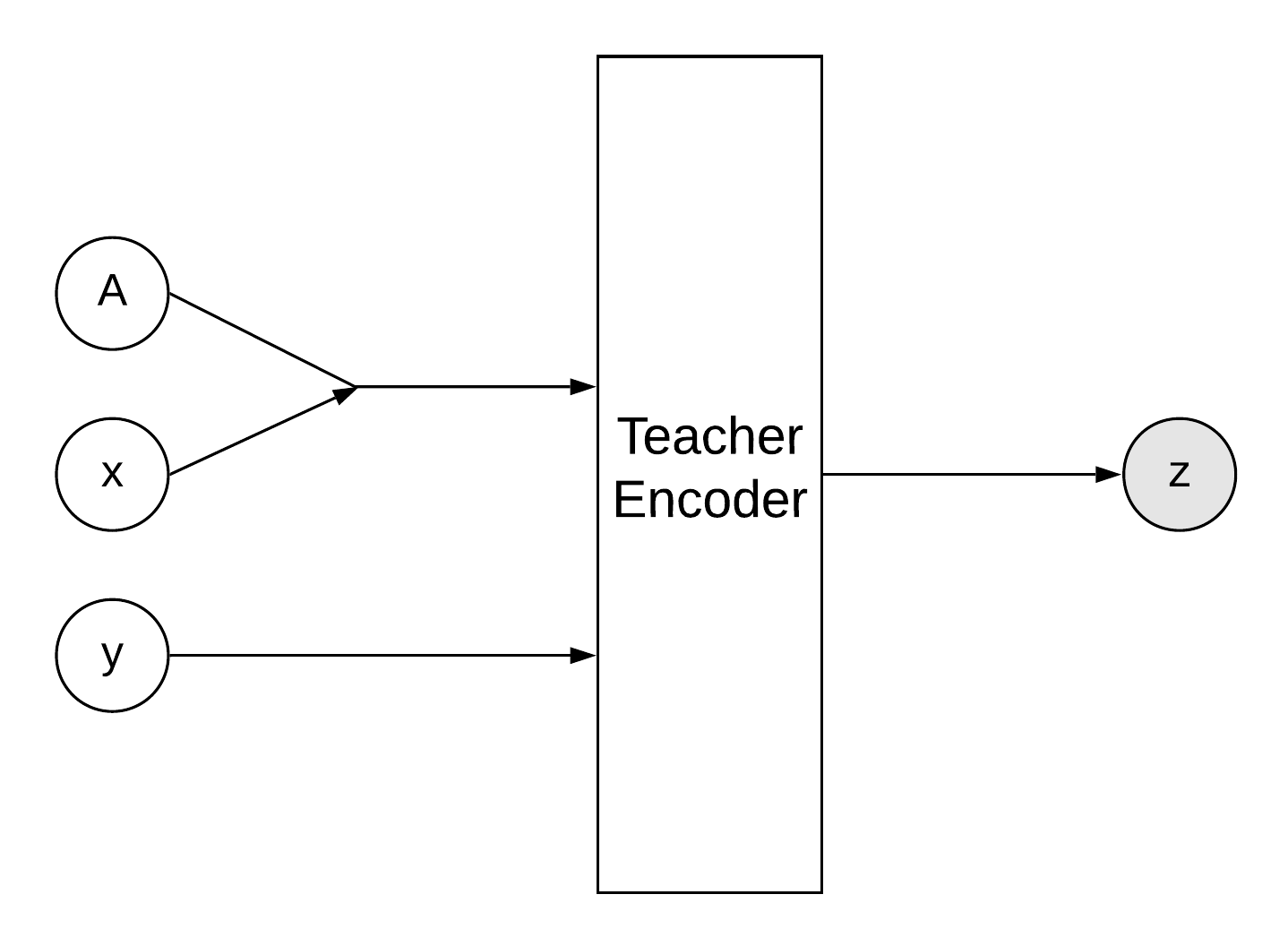} & \includegraphics[width=0.3\textwidth]{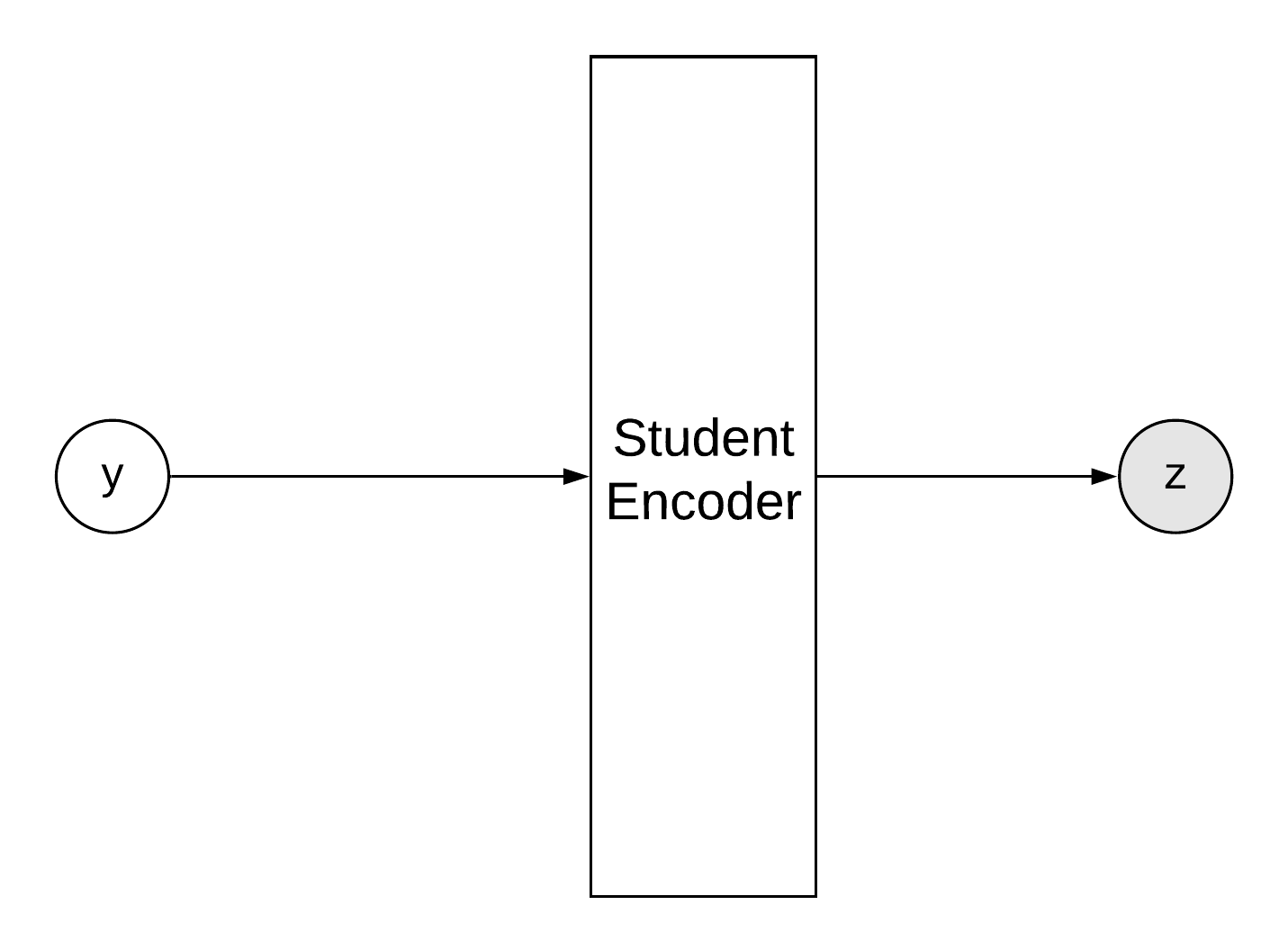}\\
 (a) teacher encoder $q_\theta(z|x,y,\mathcal{A})$ & (b) student encoder $p_{\phi_1}(z|y)$
\end{tabular}
\caption{Probabilistic encoders in the framework. Shaded nodes denote the random variable.}	
\label{fig:encoders}
\end{figure*}

The framework employs two encoders of $z$: a teacher encoder $q_\theta(z|x,y,\mathcal{A})$ and a student encoder $p_{\phi_1}(z|y)$;
see Figure \ref{fig:encoders}. The student encoder $p_{\phi_1}(z|y)$ takes the observation $y$, and encodes the observation-based knowledge so as to inform the recurrent unit $h_{\phi_2}$. Given one sample $z$ from $p_{\phi_1}(z|y)$, the network $h_{\phi_2}$ gives one refining
increment, and the distribution of $z$ contributes to the diversity of the unknown image $x$.
To help train the student encoder $p_{\phi_1}(z|y)$, we provide  the teacher encoder $q_\theta(z|x,y,\mathcal{A})$ with the ground-truth $x$ and the forward map $\mathcal{A}$.
The teacher encoder $q_\theta(z|x,y,\mathcal{A})$ is discarded once the training is finished.
The encoders $p_{\phi_1}(z|y)$ and $q_\theta(z|x,y,\mathcal{A})$ are modelled by two DNNs $h_{\phi_1}(\cdot)$ and $h_\theta(\cdot)$, respectively, which reparameterize the auxiliary variable $z$.
Since $z$ is low-dimensional, CNNs with reduced mean layers and $1\times 1$ convolutional layers can guarantee the dimension flexibility of the input $y$.
To input the ground-truth data $x$ and $y$ into $h_\theta(\cdot)$, we use the forward map $\mathcal{A}$ and concatenate $\mathcal{A}(x)$ with $y$ along the channel axis. This construction is very flexible with the problem dimension, and allows training $(x_i, y_i)$ of varying size (and the corresponding $\mathcal{A}_i$).

Now we can state the complete algorithms for training and inference of the cVAE framework for uncertainty-aware learned image reconstruction in Algorithms \ref{alg:training} and
\ref{alg:sampling}, respectively. Here $M$ denotes the mini-batch size, $T$ the maximum number of training batches, $K$ the
number of recurrences of $h_{\phi_2}$ for one sample, and $(\hat{\phi}, \hat{\theta})$ the output of the training procedure (i.e., the learned parameters). There are many possible choices of the stochastic optimizer at line 11 of Algorithm \ref{alg:training} (e.g., ADAM \cite{KingmaBa:2015}, and SGD). We shall employ ADAM \cite{KingmaBa:2015} in our experiment. The final sample from the
recurrent process is regarded as the mean of the conditional distribution $p_\phi(x|y, z,\mathcal{A})$. Thus, given the
initial $x^0$, the iteration with different realisations of $z$ leads to diverse samples of the unknown image $x$.
Since each sample is the mean of $p_\phi(x|y, z,\mathcal{A})$ rather than a direct sample from the approximate posterior $p_\phi(x|y)$, the summarizing statistics also has to be transformed; see \eqref{eq:stat}.
Note that the posterior variance contains two components: one is due to the background (i.e., $\beta I$),
and the other is due to the sample variance (i.e., $\frac{1}{S}\sum_{i=1}^S x_ix_i^t - \widehat{\mathbb{E}}_{p_\phi(x|y)}[x]\widehat{\mathbb{E}}_{p_\phi(x|y)}[x]^t$), as shown in the following result.
\begin{proposition}\label{thm:stats}
Let $p_{\phi}(x|y,z,\mathcal{A})=\mathcal{N}(x|x^K(z),\beta I)$ the approximate posterior $p_\phi(x|y)=\int p_{\phi_2}(x|y,z,\mathcal{A})p_{\phi_1}(z|y)\mathrm{d}z$
be a mixture of Gaussian distributions with $z$ being the mixture
variable. Then given samples $\{z_i\}_{i=1}^S$ of $z$ from $p_{\phi}(z|y)$, and the corresponding $x^K(z)$, denoted
by $\{x_i\}_{i=1}^S$, the mean $\mathbb{E}_{p_\phi(x|y)}[x]$ and the covariance $\mathrm{Cov}_{p_\phi(x|y)}[x]$ of
$p_\phi(x|y)$ can be estimated by
\begin{equation}\label{eq:stat}
  \begin{aligned}
	\widehat{\mathbb{E}}_{p_\phi(x|y)}[x] &= \frac{1}{S}\sum_{i=1}^S x_i,\\
	\widehat{\mathrm{Cov}}_{p_\phi(x|y)}[x] &= \beta I + \frac{1}{S}\sum_{i=1}^S x_ix_i^t - \widehat{\mathbb{E}}_{p_\phi(x|y)}[x]\widehat{\mathbb{E}}_{p_\phi(x|y)}[x]^t.
   \end{aligned}
\end{equation}
\end{proposition}
\begin{proof}
For the mean $\mathbb{E}_{p_\phi(x|y)}[x]$, by dentition, there holds
\begin{equation*}
\begin{split}
	\mathbb{E}_{p_\phi(x|y)}[x] &= \int_x x p_\phi(x|y) \text{d}x
	= \int_x x\int_z p_{\phi_2}(x|y,z,\mathcal{A})p_{\phi_1}(z|y)\text{d}z\text{d}x\\
	&= \int_z \int_x xp_{\phi_2}(x|y,z,\mathcal{A})\text{d}xp_{\phi_1}(z|y)\text{d}z
	= \int_z x^K(z)p_{\phi_1}(z|y)\text{d}z.
\end{split}
\end{equation*}
Thus, $\frac{1}{S}\sum_{i=1}^S x_i$ is an unbiased estimator of $\mathbb{E}_{p_\phi(x|y)}[x]$.
Similarly, for the covariance $\mathrm{Cov}_{p_\phi(x|y)}[x]$, by the standard bias variance decomposition,
\begin{equation*}
	\mathrm{Cov}_{p_\phi(x|y)}[x] = \int xx^T p_\phi(x|y)\text{d}x - \mathbb{E}_{p_\phi(x|y)}[x]\mathbb{E}_{p_\phi(x|y)}[x]^T
\end{equation*}
Now the first term on the right hand side, there holds
\begin{equation*}
\begin{split}
	\int xx^T p_\phi(x|y)\text{d}x &= \int_x xx^T \int_z p_{\phi_2}(x|y,z,\mathcal{A})p_{\phi_1}(z|y)\text{d}z\text{d}x\\
	&= \int_z \int_x xx^Tp_{\phi_2}(x|y,z,\mathcal{A})\text{d}xp_{\phi_1}(z|y)\text{d}z\\
	&= \int_z (\mathrm{Cov}_{p_{\phi_2}(x|y,z,\mathcal{A})}[x] + \mathbb{E}_{p_{\phi_2}(x|y,z,\mathcal{A})}[x]\mathbb{E}_{p_{\phi_2}(x|y,z,\mathcal{A})}[x]^T)p_{\phi_1}(z|y)\text{d}z\\
	&= \beta I + \int_z \mathbb{E}_{p_{\phi_2}(x|y,z,\mathcal{A})}[x]\mathbb{E}_{p_{\phi_2}(x|y,z,\mathcal{A})}[x]^Tp_{\phi_1}(z|y)\text{d}z
\end{split}
\end{equation*}
Consequently, $\beta I + \frac{1}{S}\sum_{i=1}^S x_ix_i^t - \widehat{\mathbb{E}}_{p_\phi(x|y)}[x]\widehat{\mathbb{E}}_{p_\phi(x|y)}[x]^t$
is an unbiased estimator of the covariance $\mathrm{Cov}_{p_\phi(x|y)}[x]$.
\end{proof}

\begin{algorithm}[hbt!]
\centering
\caption{Training procedure}\label{alg:training}
\begin{algorithmic}[1]
	\STATE Input: Training data $\{(x_i, y_i,\mathcal{A}_i)\}_{i=1}^N$, $\beta$, $T$, $K$, $M$
	\FOR{$t=1,2,\ldots,T$}
	\STATE Randomly select a mini-batch training data $\{( x_i, y_i,\mathcal{A}_i)\}_{i=1}^M$;
	\STATE Sample $\{z_i\}_{i=1}^M$ from $\{q_\theta(z|x_i, y_i)\}_{i=1}^M$;
	\STATE Initialize $\{\hat{x}_i\}_{i=1}^M$ with $\{\mathcal{A}_i^*(y_i)\}_{i=1}^M$ and $\{a_i\}_{i=1}^M$ with zeros;
	\FOR{$k=1,2,\ldots,K$}
	\STATE Update $\{\hat{x}_i\}_{i=1}^M$ and  $\{a_i\}_{i=1}^M$ with $\{h_{\phi_2}(\hat{x}_i, E(\mathcal{A}_i, y_i,\hat{x}_i), R(\hat{x}_i), a_i, z_i)\}_{i=1}^M$;
	\ENDFOR
    \STATE Evaluate the KL divergence $\{D_{\rm KL}(q_\theta(z|x_i,y_i)||p_{\phi_1}(z|y_i))\}_{i=1}^M$;
    \STATE Compute the objective function $\mathcal{\hat L}(\phi, \theta)$;
    \STATE Update the parameters $(\phi, \theta)$;
	\ENDFOR
    \STATE Output: $(\hat{\phi}, \hat{\theta})$
\end{algorithmic}
\end{algorithm}

\begin{algorithm}[hbt!]
\centering
\caption{Inference procedure}\label{alg:sampling}
\begin{algorithmic}[1]
	\STATE Input: Test data $(\mathcal{A}, y)$, $S$, $K$, $\hat{\phi}=(\hat{\phi_1}, \hat{\phi}_2)$
	\FOR{$s=1,2,\ldots,S$}
	\STATE Sample $z_s$ from $p_{\phi_1}(z|y)$
	\STATE Initialise $\hat{x}_s$ with $\mathcal{A}^*(y)$ and $a$ with zeros
	\FOR{$k=1,2,\ldots,K$}
	\STATE Update $\hat{x}_s$ and $a$ with $h_{\phi_2}(\hat{x}_s, E(\mathcal{A}, y, \hat{x}_s), R(\hat{x}_s), a, z_s)$
	\ENDFOR
	\ENDFOR
    \STATE Output: $\{\hat{x}_s\}_{s=1}^S$
    \STATE Evaluate $(\widehat{\mathbb{E}}_{p(x|y)}[x], \widehat{\mathrm{Cov}}_{p(x|y)}[x])$ by Eq. \eqref{eq:stat}
\end{algorithmic}
\end{algorithm}

\section{Numerical experiments and discussions}\label{sec:numerics}

Now we showcase the proposed cVAE framework for learned reconstruction with quantified aleatoric uncertainty with numerical experiments on positron emission tomography (PET).
PET is a pillar of modern diagnostic imaging, allowing noninvasive, sensitive and specific detection of functional changes in a number of diseases. Most conventional PET reconstruction algorithms rely on penalized maximum likelihood estimates, using a hand crafted prior (e.g., total variation and anatomical); see the work \cite{QiLeahy:2006} for an overview on classical reconstruction techniques. More recently learning based approaches have been proposed. While these techniques have been successful, they still lack the capability to provide uncertainty estimates; see the work \cite{arridge2018variational, FilipovicBarat:2019, zhang2019expectation, ZhouYuZhang:2020} for several recent investigations on UQ in PET reconstruction, but none of which is based on deep learning.

\begin{figure*}[htp!]
\centering
\includegraphics[width=0.9\textwidth]{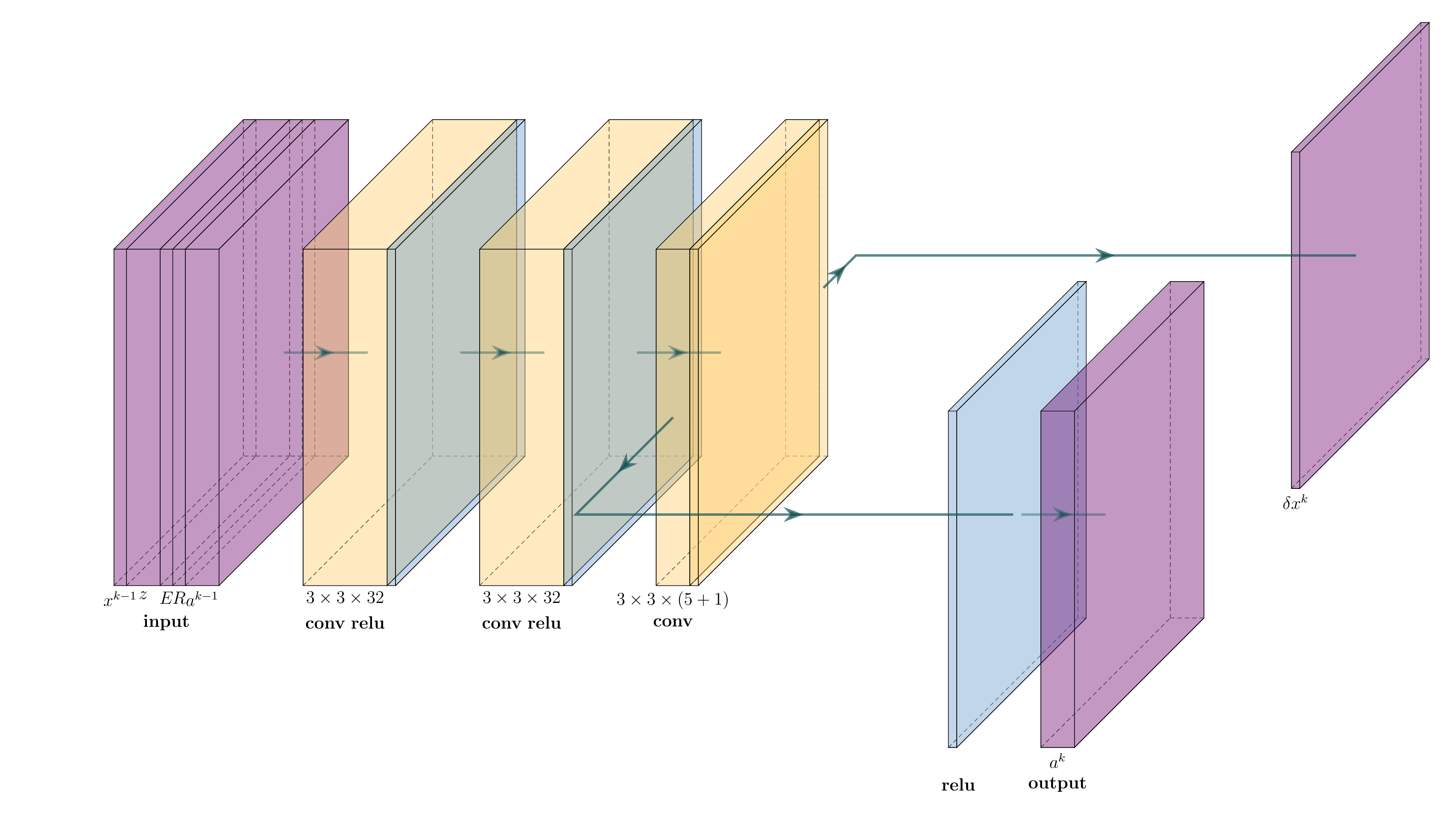}
\caption{The layer configuration of the recurrent unit $h_{\phi_2}$: $3\times 3\times 32$
denotes convolutional layer with a kernel size $3\times 3$ and $32$ output channels. In the third convolutional layer,
$5+1$ denotes $5$ channels for memory $a^k$ and $1$ channel for the update $\delta^k$.}	
\label{fig:iter_net_layers}
\end{figure*}

\begin{figure*}[htp!]
\centering
\begin{tabular}{c}
\includegraphics[width=0.9\textwidth]{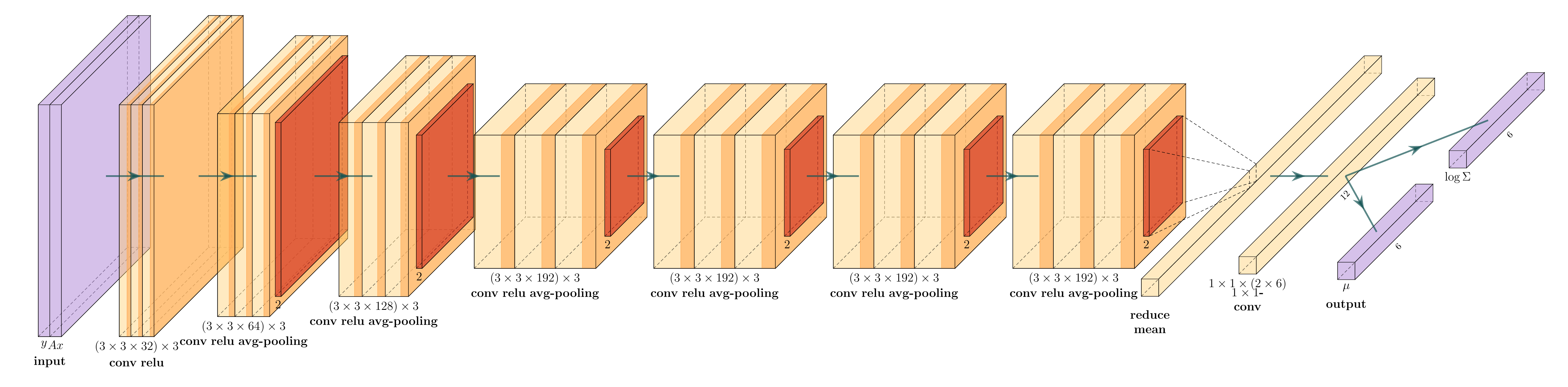}\\
\includegraphics[width=0.9\textwidth]{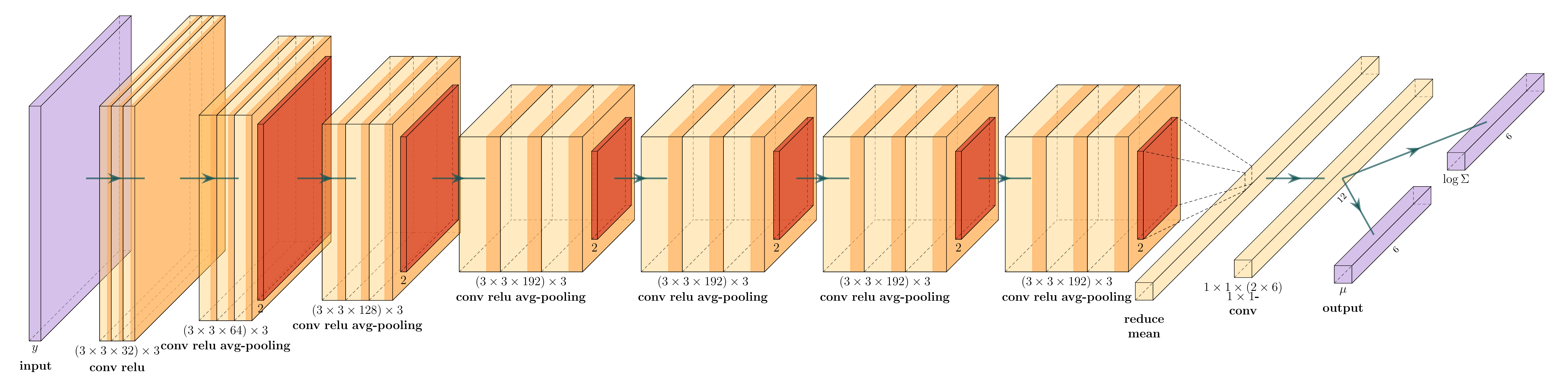}
\end{tabular}
\caption{The layer configurations of the teacher (top) and student (bottom) encoder:
$(3\times 3\times 32)\times 3$ denotes $3$ convolutional layers respectively followed by an ReLU layer with a kernel size $3\times 3$
and $32$ output channels. $2$ under the brown layer denote average pooling layer with stride size $2$. $1\times 1\times
(2\times 6)$ denotes $1\times 1$ convolutional layer with $12$ output channels, i.e. $6$ for mean $\mu$ and $6$ for
log (diagonal) variance $\log\Sigma$.}	
\label{fig:encoder_layers}
\end{figure*}

For the experiments below, we employ a 3-layer CNN as the recurrent unit $h_{\phi_2}$ and fix $K=10$ iterations for each
sampling step, cf. Figure \ref{fig:iter_net_layers} for a schematic illustration, and VGG style encoders for both $h_{\phi_1}$ and $h_\theta$, cf. Figure \ref{fig:encoder_layers}.
We train the network on a synthetic dataset consisting of elliptical phantoms, and test it on the public medical imaging dataset BrainWeb \cite{cocosco1997brainweb} (available from \url{https://brainweb.bic.mni.mcgill.ca/}).
Throughout, the training pair $(x,y)\in \mathbb{R}^{128\times 128} \times \mathbb{R}^{30\times 183}$, and the forward map is the Radon transform, which is normalised, and different peak values of $x$ are used to indicate the count level: ${\rm 1e4}$ and ${\rm 1e2}$ for respectively moderate and low count levels.
The observation $y$ is generated by corrupting the sinogram $\mathcal{A}x$ by Poisson noise entrywise, i.e., $y_i\sim \mathrm{Pois}((\mathcal{A}x)_i)$, where $\mathrm{Pois}(\cdot)$ denotes the Poisson random variable.
The hyper-parameter $\beta$ is tuned in a trial-and-error manner, and fixed at $\beta=\text{5e-3}$ below.
The experiments are conducted on a desktop with two Nvidia GeForce 1080 Ti GPUs and Intel i7-7700K CPU 4.20GHz$\times$8.
It is trained for $T={\rm 1e5}$ batches, each of which contains 10 randomly generated $(x,y)$ pairs on-the-fly.
The training almost converges after $\rm 2e4$ batches and it takes around $11$ hours to go over all $T={\rm1e5}$ batches.
The summarizing statistics reported below are computed from $1000$ samples for each observation $y$ generated by the trained network.
The implementation uses the following public deep learning frameworks: \texttt{Tensorflow} (\url{https://www.tensorflow.org/}, \cite{abadi2016tensorflow}),
\texttt{Tensorflow Probability} (\url{https://www.tensorflow.org/probability}, \cite{dillon2017tensorflow}), \texttt{DeepMind Sonnet} (\url{github.com/deepmind/sonnet})
and \texttt{ODL} (\url{github.com/odlgroup/odl}), and the source code will be made available at
\url{github.com/chenzxyz/cvae}.

First we compare the proposed cVAE approach with conventional and deep learning based methods on all 181 phantoms in the BrainWeb dataset. It is compared with the following three benchmark methods:  maximum likelihood EM (MLEM) \cite{shepp1982maximum},
total variation (TV) \cite{RudinOsherFatemi:1992} with nonnegativity constraint and learned gradient descent (LGD) \cite{adler2017solving}.
MLEM and TV are two established reconstruction methods in the PET community, and LGD is an unrolled iterative method
inspired by classical variational regularization and exploits DNNs for iterative refinement.
For MLEM, we use the \texttt{odl} inbuilt solver \texttt{mlem}, and for TV, use the primal dual hybrid gradient method (implemented by \texttt{odl.solvers.phgd}).
The regularization parameter $\alpha$ for total variation prior is fixed at $\alpha=\text{2e-1}$ and $\alpha={\rm 2e0}$ for the moderate and low count levels, respectively, which is determined in a trial-and-error manner.
The comparative results are summarized in Table \ref{tab:Compare-quant}, shown with two most popular image quality metrics, i.e., SSIM and PSNR, averaged over all $181$ phantoms in the BrainWeb dataset. The results clearly show that cVAE can deliver state-of-the-art point estimates in terms of PSNR and SSIM, especially in the practically very
important low count case.
Compared with these deterministic benchmark methods, cVAE can additionally provide uncertainty information.

\begin{table}[hbt!]
\centering
\caption{Comparisons between cVAE mean and benchmark methods on 181 BrainWeb phantoms at two count levels: ${\rm 1e4}$ (MC)
and ${\rm 1e2}$ (LC). The numbers in the table denote the SSIM/PSNR values. \label{tab:Compare-quant}}
\begin{tabular}{lcccc}
 \hline
   & MLEM         &  TV      &  LGD       &  cVAE\\
 \hline\hline
MC &  0.74/23.20  &  0.85/\textbf{28.76}  &  \textbf{0.92}/\textbf{29.07}  &  \textbf{0.91}/28.01\\
 \hline
LC &  \textbf{0.64}/21.55  &  0.62/\textbf{22.58}  &  0.59/21.68  &  \textbf{0.64}/\textbf{23.10}\\
 \hline
\end{tabular}
\end{table}

Next we compare cVAE with a probabilistic approach \cite{lakshminarayanan2017simple}, which reports state-of-the-art performance for aleatoric uncertainty (see \cite{HeTeh:2020} for theoretical interpretations). It employs (non-Bayesian) neural network ensembles to estimate predictive uncertainty, where each network in the ensemble learns similar values close to the training data, and different ones in regions of the space far from the training data. For the comparison, we train a mixture with three multivariate Gaussians (GM3) without adversarial samples, where the training of each component of the mixture is to fit a mean network and a variance network using Gaussian log-likelihood to the outputs \cite{nix1994estimating}.
To stabilize the training procedure, we first train the mean network, and then train the variance network. Alternatively, one can train the mean network as a warm up and then train
the mean and variance networks simultaneously, but it usually leads to worse results, and thus we do not present the corresponding results.
The comparative quantitative results are given in Table \ref{tab:Compare_prob_quant}, which present the PSNR and SSIM results for ten phantoms from the Brainweb dataset, in order to shed fine-grained insights into the performance of the methods over different phantoms. It is observed that in the low count case, cVAE can provide better point estimates in terms of both SSIM and PSNR, which concurs with Figure \ref{fig:prob_bench}, whereas in the moderate count case, GM3 can sometimes deliver better results.
In terms of the variance map, the results by GM3 contain more structural patterns, and resemble more closely the error map.

\begin{table}[hbt!]
\centering
\setlength{\tabcolsep}{4.2pt}
\caption{PSNR and SSIM values for the reconstructions by the trained cVAE  and GM3 on ten phantoms
with peak value ${\rm1e4}$ (MC) and $\rm 1e2$ (LC). The column index refers to \texttt{Python} style
index of the phantom in the BrainWeb dataset. The PSNR and SSIM values are shown as cVAE/GM3.
\label{tab:Compare_prob_quant}}
\begin{tabular}{ccccccc}
\toprule
 &  & 10          & 20          & 30          & 50          & 70         \\
\midrule
\multirow{ 2}{*}{PNSR} & MC & 27.66/28.05 & 27.14/27.48 & 27.25/27.43 & 27.25/27.50 & 25.65/26.77 \\
                    & LC & 22.60/21.86 & 22.09/21.35 & 22.30/21.09 & 22.14/20.69 & 20.87/19.32 \\
\midrule
\multirow{ 2}{*}{SSIM} & MC & 0.89/0.89   & 0.89/0.89   & 0.91/0.91   & 0.88/0.88   & 0.88/0.89   \\
                    & LC & 0.65/0.62   & 0.65/0.61   & 0.69/0.62   & 0.54/0.50   & 0.46/0.45   \\
\midrule
 &  & 90           & 100         & 110         & 130         & 150\\
 \midrule
\multirow{ 2}{*}{PSNR} & MC & 24.98/26.83 & 26.91/27.74 & 27.81/28.43 & 27.96/29.33 & 30.86/32.57\\
                    & LC & 20.52/19.02 & 21.39/19.74 & 22.22/20.61 & 22.48/21.32 & 25.78/23.67\\
 \midrule
\multirow{ 2}{*}{SSIM} & MC & 0.88/0.91   & 0.90/0.91   & 0.92/0.92   & 0.94/0.94   & 0.96/0.96\\
                    & LC & 0.60/0.53   & 0.62/0.55   & 0.61/0.57   & 0.60/0.57   & 0.73/0.67\\
\bottomrule
\end{tabular}
\end{table}

\newcommand{\myfigwidtht}{0.24\linewidth}
\begin{figure}[htb!]
\centering
\setlength{\tabcolsep}{1pt}
\begin{tabular}{cccccccccc}
\toprule
\includegraphics[width=\myfigwidtht]{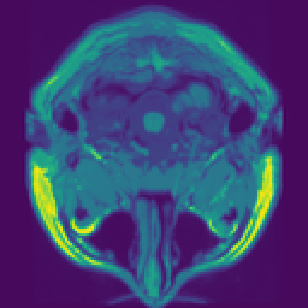} &
\includegraphics[width=\myfigwidtht]{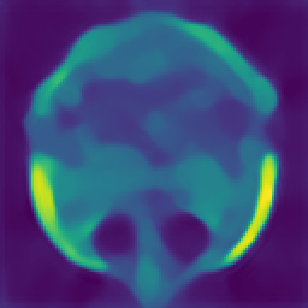} &
\includegraphics[width=\myfigwidtht]{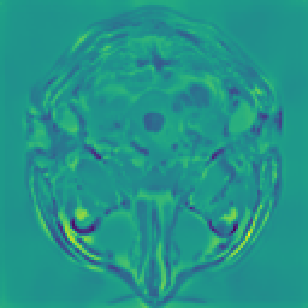} &
\includegraphics[width=\myfigwidtht]{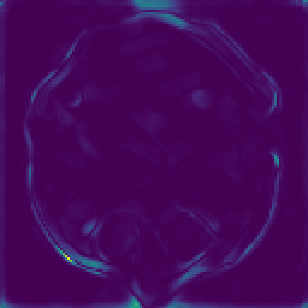} \\
$x^\dag$ & $\hat x_{\rm cvae}$ & $\hat x_{\rm cvae}-x^\dag$ & $\mathrm{Var}(x_{\rm cvae})$\\
&\includegraphics[width=\myfigwidtht]{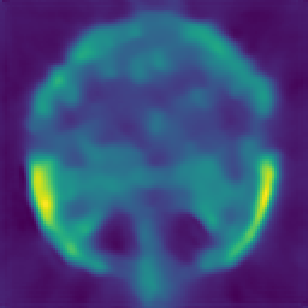} &
\includegraphics[width=\myfigwidtht]{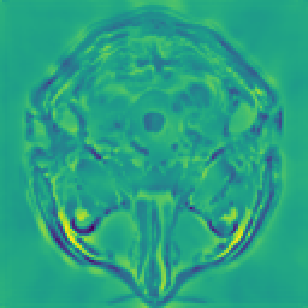} &
\includegraphics[width=\myfigwidtht]{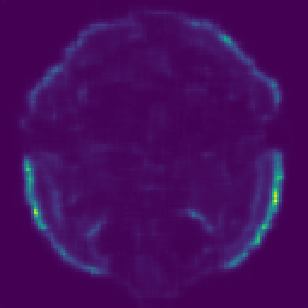} \\
& $\hat x_{\rm gm3}$ & $\hat x_{\rm gm3}-x^\dag$ & $\mathrm{Var}(x_{\rm gm3})$\\
\midrule
\includegraphics[width=\myfigwidtht]{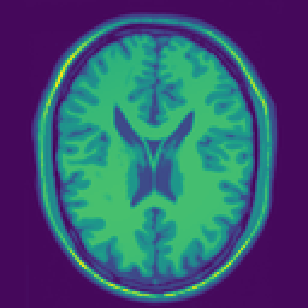} &
\includegraphics[width=\myfigwidtht]{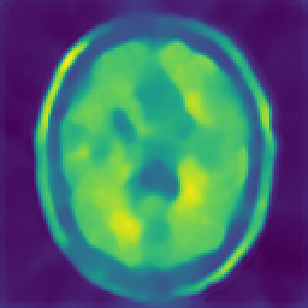} &
\includegraphics[width=\myfigwidtht]{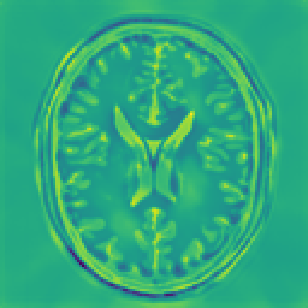} &
\includegraphics[width=\myfigwidtht]{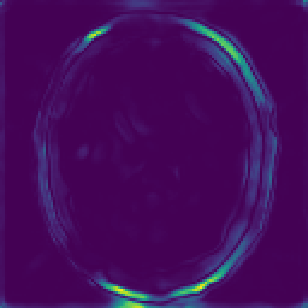} \\
$x^\dag$ & $\hat x_{\rm cvae}$ & $\hat x_{\rm cvae}-x^\dag$ & $\mathrm{Var}(x_{\rm cvae})$\\
&\includegraphics[width=\myfigwidtht]{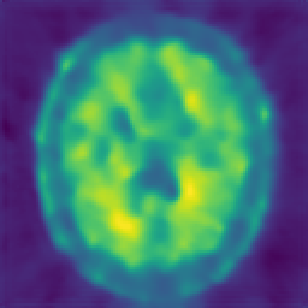} &
\includegraphics[width=\myfigwidtht]{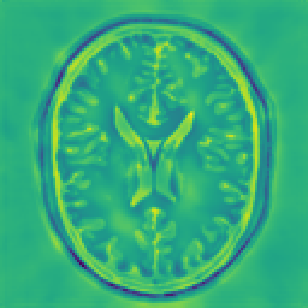} &
\includegraphics[width=\myfigwidtht]{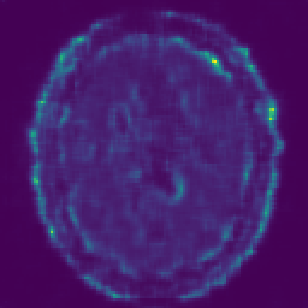} \\
& $\hat x_{\rm gm3}$ & $\hat x_{\rm gm3}-x^\dag$ & $\mathrm{Var}(x_{\rm gm3})$\\
\bottomrule
\end{tabular}
\caption{Reconstructions of two phantoms (i.e., 10 for the two rows and 90 for the bottom two rows) from BrainWeb with peak value $\rm 1e2$, by cVAE and GM3,
respectively, in terms of the mean $\hat x$, the pointwise error $\hat x-x^dag$ and the posterior variance $\mathrm{Var}(x)$.}	
\label{fig:prob_bench}
\end{figure}

\begin{figure}[htb!]
\centering
\setlength{\tabcolsep}{2pt}
\begin{tabular}{ccc}
\toprule
\multicolumn{3}{c}{MC, Phantom 90} \\
\includegraphics[width=0.30\linewidth]{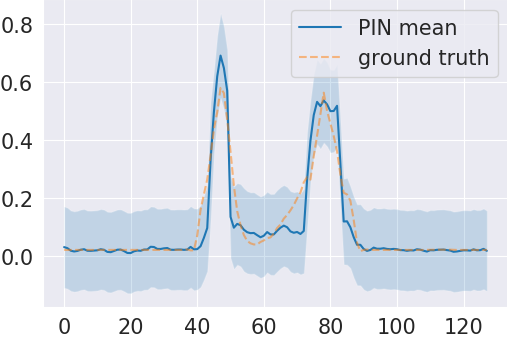} &
\includegraphics[width=0.30\linewidth]{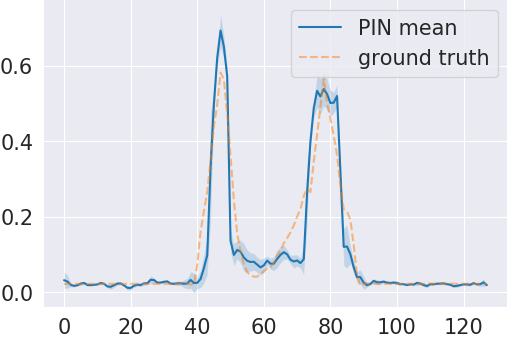} &
\includegraphics[width=0.30\linewidth]{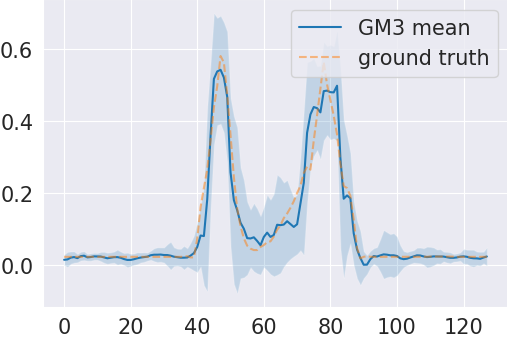}      \\
\includegraphics[width=0.30\linewidth]{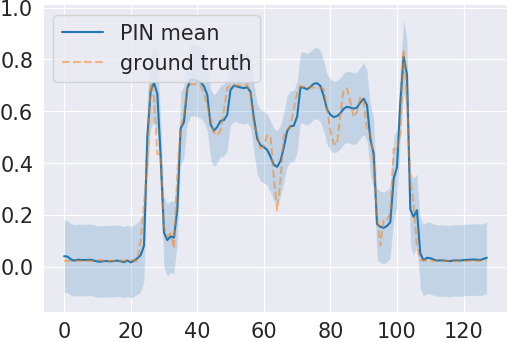} &
\includegraphics[width=0.30\linewidth]{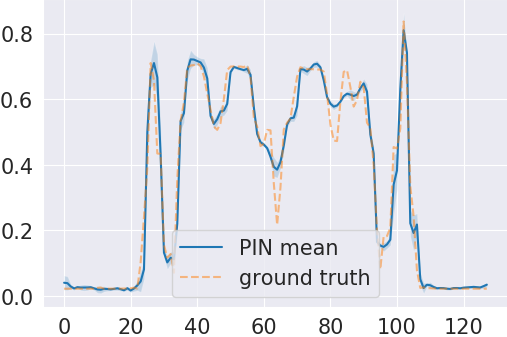} &
\includegraphics[width=0.30\linewidth]{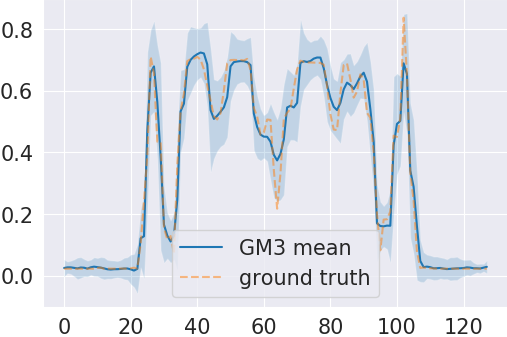}      \\
\midrule
\multicolumn{3}{c}{LC, Phantom 90}\\
\includegraphics[width=0.30\linewidth]{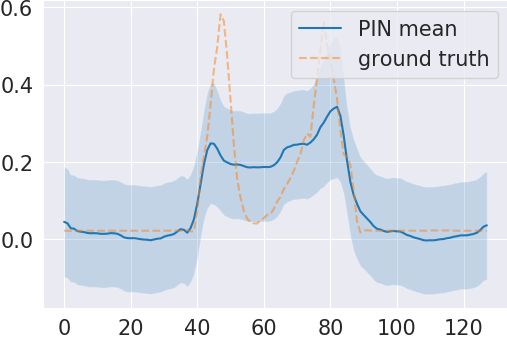} &
\includegraphics[width=0.30\linewidth]{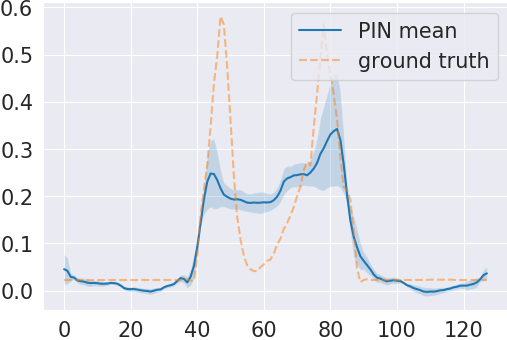} &
\includegraphics[width=0.30\linewidth]{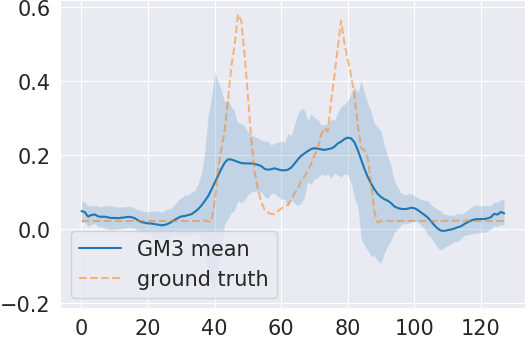} \\
\includegraphics[width=0.30\linewidth]{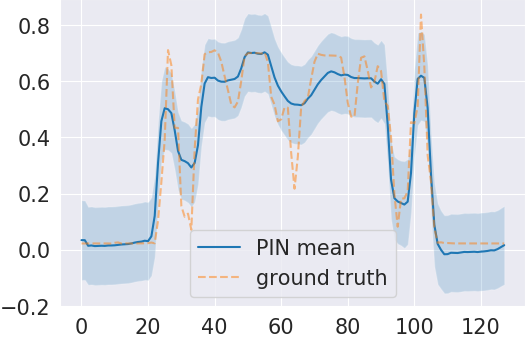} &
\includegraphics[width=0.30\linewidth]{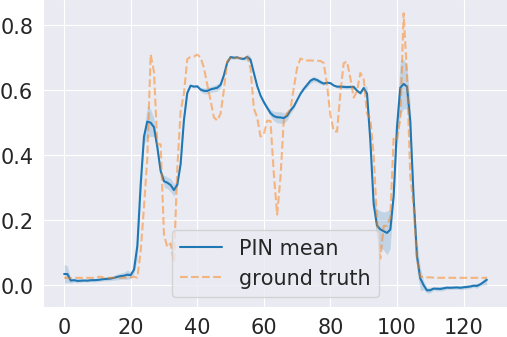} &
\includegraphics[width=0.30\linewidth]{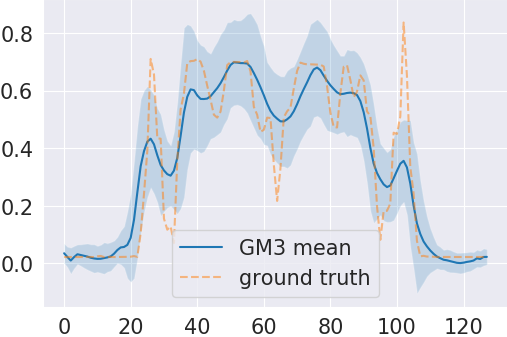}\\
cVAE-FV & cVAE-WB & GM3-FV\\ 
\bottomrule
\end{tabular}
\caption{Comparison between cVAE with full variance (cVAE-FV), cVAE without background variance (cVAE-WB) and GM3 with full variance (GM3-FV),
for BrainWeb phantom 90 (size: $128\times 128$) with the two peak values ${\rm 1e4}$ (MC) and ${\rm 1e2}$ (LC). Within each block,
from left to right: sample mean and 0.95 credible interval of the 11th (top) and 101-th (bottom) horizontal slice.}	
\label{fig:interval}
\end{figure}

To shed more insights into the variance by cVAE and the benchmark GM3, we show the cross-section plots with $0.95$ Highest Posterior Density (HPD) in Figure \ref{fig:interval} for both moderate and low count levels. According to
Proposition \ref{thm:stats}, the estimated covariance by cVAE contains two distinct sources, i.e., sample variance and the variance $\beta$ of the conditional Gaussians $p_{\phi}(x|y,z,\mathcal{A})=\mathcal{N}(x|x^K(z),\beta I)$.
The latter is uniform across the pixels, and acts as a background.
We show the HPDs of cVAE with full variance (unbiased variance estimated by $\widehat{\mathrm{Cov}}_{p_\phi(x|y)}[x]$) and the variance without $\beta$ factor (i.e., $\widehat{\mathrm{Cov}}_{p_\phi(x|y)}[x]-\beta I$).
It is observed that the latter contains more structures in the credible intervals.
Further, the overall shape and magnitude of the HPDs by cVAE with the full variance and GM3 are fairly closely to each other.
It is noteworthy that in the cold regions (i.e. zero count), cVAE can provide almost zero variance, upon subtracting the background variance, and is able to indicate the contrast of variance to highlight the pixels, whereas the variances by GM3 are also relatively high.
The comparison between the cross-section plots for low and moderate count cases (i.e.,
high and low noise levels) on the same ground-truth phantom indicates that cVAE does provide higher uncertainty for higher noise level, which is intuitively consistent with the underlying statistical background.

\begin{figure}[htb!]
\centering
\setlength{\tabcolsep}{1.5pt}
\begin{tabular}{cccccccc}
\includegraphics[width=\myfigwidtht]{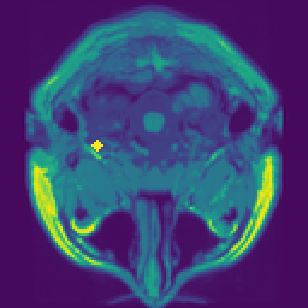} &
\includegraphics[width=\myfigwidtht]{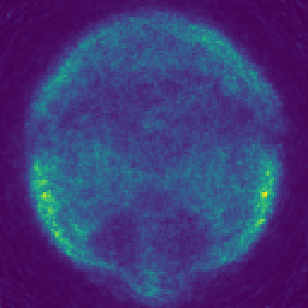} &
\includegraphics[width=\myfigwidtht]{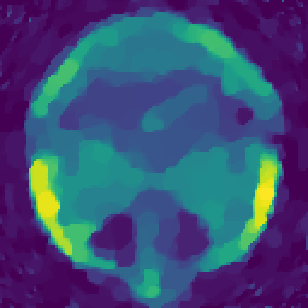} &
\includegraphics[width=\myfigwidtht]{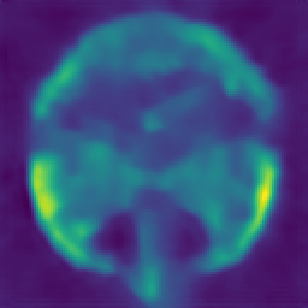} \\
$x^\dag$ & $x_{\rm mlem}$ & $x_{\rm tv}$ & $x_{ldg}$ \\
\includegraphics[width=\myfigwidtht]{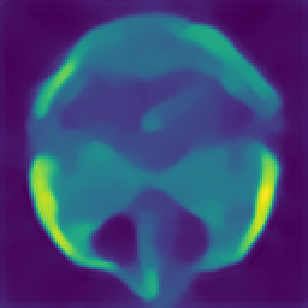} &
\includegraphics[width=\myfigwidtht]{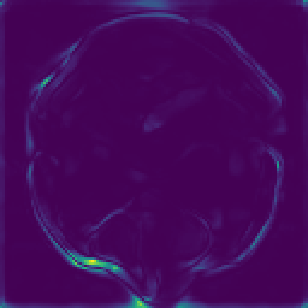} &
\includegraphics[width=\myfigwidtht]{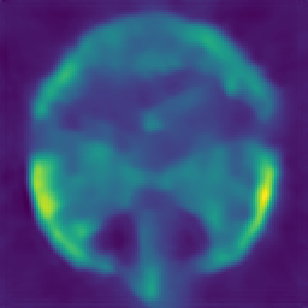} &
\includegraphics[width=\myfigwidtht]{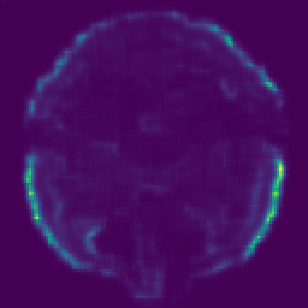}\\
$\hat x_{\rm cvae}$ & $\mathrm{Var}(x_{\rm cvae})$ & $\hat x_{\rm gm3}$ & $\mathrm{Var}(x_{\rm gm3}) $\\
\includegraphics[width=\myfigwidtht]{tumour_lc_010_gt.png}&
\includegraphics[width=\myfigwidtht]{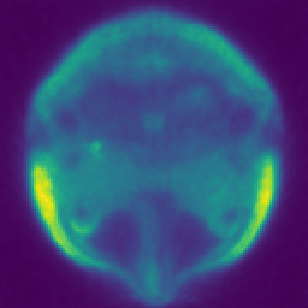} &
\includegraphics[width=\myfigwidtht]{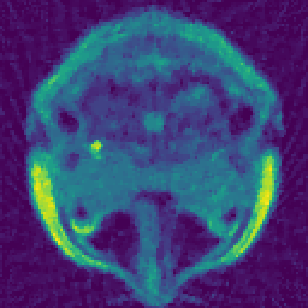} &
\includegraphics[width=\myfigwidtht]{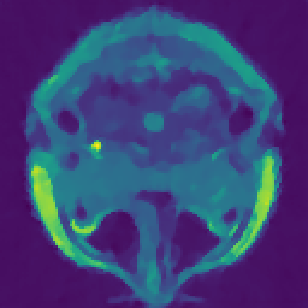} \\
$x^\dag$ & $x_{\rm mlem}$ & $x_{\rm tv}$ & $x_{ldg}$ \\
\includegraphics[width=\myfigwidtht]{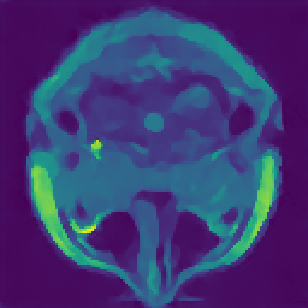} &
\includegraphics[width=\myfigwidtht]{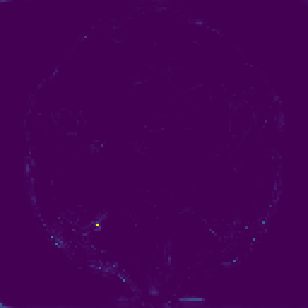} &
\includegraphics[width=\myfigwidtht]{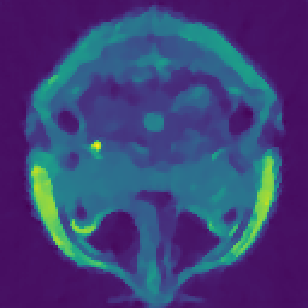} &
\includegraphics[width=\myfigwidtht]{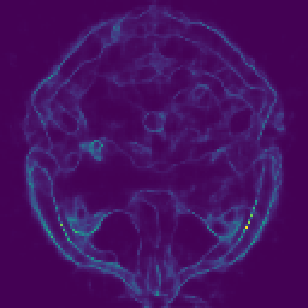}\\
$\hat x_{\rm cvae}$ & $\mathrm{Var}(x_{\rm cvae})$ & $\hat x_{\rm gm3}$ & $\mathrm{Var}(x_{\rm gm3}) $\\
\end{tabular}
\caption{Reconstructions for one BrainWeb phantom (No. 10) with tumor, obtained by benchmark methods
(MLEM, TV, LGD, GM3) and cVAE. For each phantom, the top two row is for the low count level and the
bottom two rows for the moderate count level.}\label{fig:tumour}
\end{figure}

\begin{figure}[htb!]
\centering
\setlength{\tabcolsep}{1.5pt}
\begin{tabular}{cccccccc}
\toprule
\includegraphics[width=\myfigwidtht]{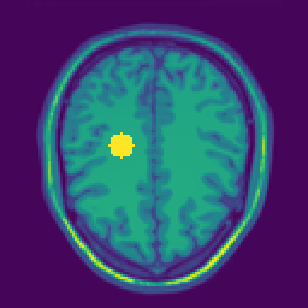} &
\includegraphics[width=\myfigwidtht]{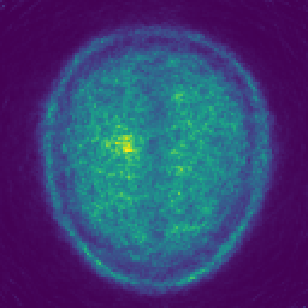} &
\includegraphics[width=\myfigwidtht]{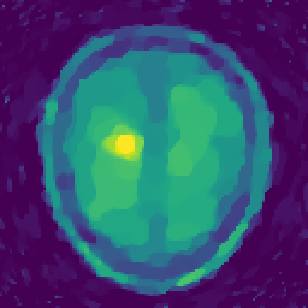} &
\includegraphics[width=\myfigwidtht]{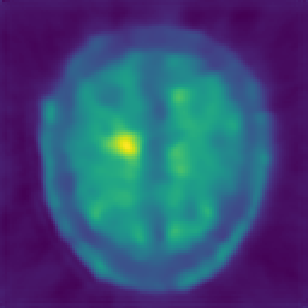} \\
$x^\dag$ & $x_{\rm mlem}$ & $x_{\rm tv}$ & $x_{ldg}$ \\
\includegraphics[width=\myfigwidtht]{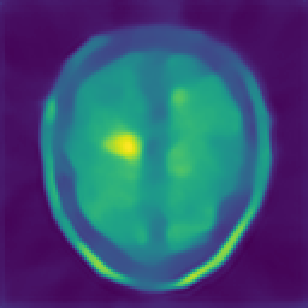} &
\includegraphics[width=\myfigwidtht]{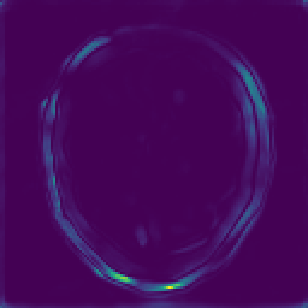} &
\includegraphics[width=\myfigwidtht]{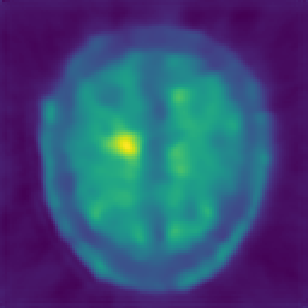} &
\includegraphics[width=\myfigwidtht]{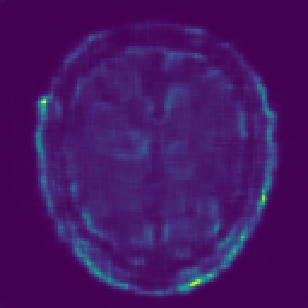}\\
$\hat x_{\rm cvae}$ & $\mathrm{Var}(x_{\rm cvae})$ & $\hat x_{\rm gm3}$ & $\mathrm{Var}(x_{\rm gm3}) $\\
\midrule
\includegraphics[width=\myfigwidtht]{tumour_lc_110_gt.png} &
\includegraphics[width=\myfigwidtht]{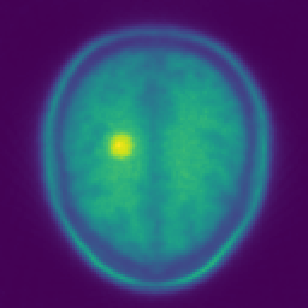} &
\includegraphics[width=\myfigwidtht]{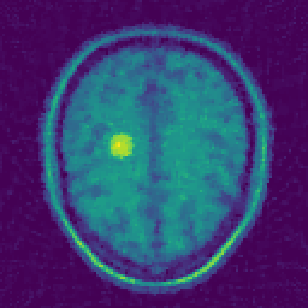} &
\includegraphics[width=\myfigwidtht]{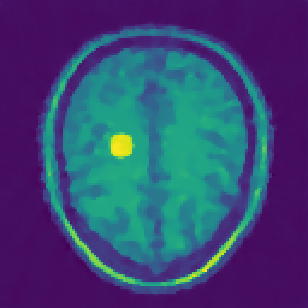} \\
$x^\dag$ & $x_{\rm mlem}$ & $x_{\rm tv}$ & $x_{ldg}$ \\
\includegraphics[width=\myfigwidtht]{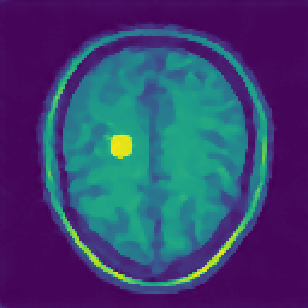} &
\includegraphics[width=\myfigwidtht]{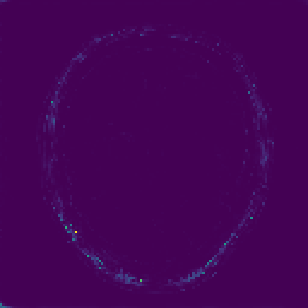} &
\includegraphics[width=\myfigwidtht]{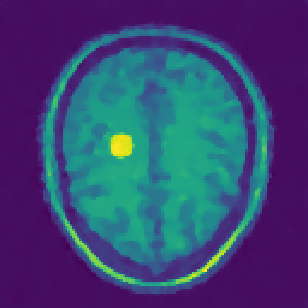} &
\includegraphics[width=\myfigwidtht]{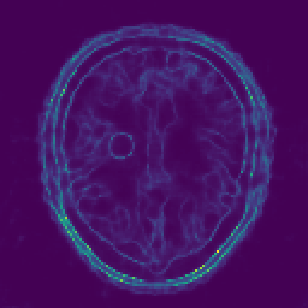}\\
$\hat x_{\rm cvae}$ & $\mathrm{Var}(x_{\rm cvae})$ & $\hat x_{\rm gm3}$ & $\mathrm{Var}(x_{\rm gm3}) $\\
\bottomrule
\end{tabular}
\caption{Reconstructions for one BrainWeb phantom (No. 110) with tumor, obtained by benchmark
methods (MLEM, TV, LGD, GM3) and cVAE. For each phantom, the top two
rows are for the low count level and the bottom two rows for the moderate count level.}\label{fig:tumour-2}
\end{figure}

Lastly, we evaluate all the methods on phantoms with an artificially added tumour by changing the pixel values to the peak value, in order to examine their capability of recovering novel features that are not presented in the training data. We (randomly) choose two phantoms from BrainWeb dataset (\texttt{Python} style index: $10$ and $110$).
A small tumour of radius $2$ and a large tumour of radius $5$ are added into the $10$-th
and $110$-th phantoms, respectively.
The corresponding reconstructions are shown in Figures \ref{fig:tumour} and \ref{fig:tumour-2}, respectively.
It is observed the tumours can be clearly reconstructed by the cVAE means for both count levels, except the small tumour at low count levels.
In the latter case, none of the methods can reasonably reconstruct the tumor, since
the data is very noisy in view of low signal strength.
The results by cVAE, LGD and GM3 are comparable, at least visually. The ability of reconstructing tumours further indicates that cVAE does not miss
out important features non-present in the training data, indicating a certain degree of robustness of the cVAE framework, so long as the signal is sufficiently strong, since many deep learning based methods tend to miss the tumour due to the bias induced by tumour-free training data \cite{MoellerMollenhoff:2019}.

\section{Conclusion}\label{sec:conclusion}

In this work, we have developed a general and flexible probabilistic computational framework
for the uncertainty quantification of inverse problems in a purely data-driven setting.
The approach is based on the conditional variational autoencoder loss, and employs the deep neural network as a recurrent unit to recurrently refine the samples using the observation and forward map, seeded by a probabilistic encoder conditioned on the observation.
The efficiency of the framework is underpinned by encoding the observations in a low-dimensional latent space.
The significant potentials of the framework have been demonstrated on PET image reconstruction with both moderate and low count levels, and the approach shows competitive performance when compared with several deterministic and probabilistic benchmark methods, especially within the
low count regime.

There are several avenues for further study. First, the framework is flexible and general, and it is of much interest to evaluate its potentials on other computationally expensive imaging modalities (e.g., MRI, CT and PET-MRI) especially in the under sampling and low-dose regime, for which there is a great demand on uncertainty quantification due to lack of sufficient information.
Such studies will also shed important insights into statistical features of the framework. Second, it is of much interest to analyze the theoretical properties of the cVAE loss as an upper bound of the expected KL divergence (e.g., approximation error and asymptotic convergence).
This line of research has been long outstanding in approximate inference, and often provides
theoretical guarantees of the overall inference procedure and guidelines for constructing efficient approximations.
Third and last, it is imperative to develop scalable benchmarks for uncertainty quantification of high-dimensional inverse problems.
Several deep learning based uncertainty quantification techniques have been proposed in the machine learning literature, but mostly on different types of uncertainties or without explicitly elucidating the sources of uncertainties (see the work \cite{BarbanoKereta:2020} for a preliminary study in medical imaging). A careful calibration of the obtained uncertainties is essential towards practical deployment, as recently highlighted in the review \cite{BarbanoArridgeJinTanno:2022}.

\bibliographystyle{abbrv}
\bibliography{cvae}
\end{document}